\documentclass[10pt]{article} 
\usepackage[accepted]{tmlr}

\usepackage{amsmath,amsfonts,bm}

\def\eqref#1{equation~\ref{#1}}

\def\1{\bm{1}}

\DeclareMathAlphabet{\mathsfit}{\encodingdefault}{\sfdefault}{m}{sl}
\SetMathAlphabet{\mathsfit}{bold}{\encodingdefault}{\sfdefault}{bx}{n}

\usepackage{color}
\usepackage{epsfig}
\usepackage{graphicx}

\usepackage{adjustbox}
\usepackage{array}
\usepackage{booktabs}
\usepackage{colortbl}
\usepackage{wrapfig}
\usepackage{hhline}
\usepackage{multirow}
\usepackage{wrapfig}
\usepackage{floatflt}

\usepackage{bm}
\usepackage{nicefrac}
\usepackage{microtype}

\usepackage{changepage}
\usepackage{extramarks}
\usepackage{fancyhdr}
\usepackage{lastpage}
\usepackage{setspace}
\usepackage{soul}
\usepackage{xspace}

\usepackage{enumerate}
\usepackage{enumitem}  

\usepackage{makecell}

\usepackage{pifont} 

\usepackage{algorithm,algpseudocode}

\usepackage[symbol]{footmisc}

\usepackage[most]{tcolorbox}

\newcolumntype{L}[1]{>{\raggedright\let\newline\\\arraybackslash\hspace{0pt}}m{#1}}
\newcolumntype{C}[1]{>{\centering\let\newline\\\arraybackslash\hspace{0pt}}m{#1}}
\newcolumntype{R}[1]{>{\raggedleft\let\newline\\\arraybackslash\hspace{0pt}}m{#1}}

\newcommand{\ignore}[1]{}

\makeatletter
\DeclareRobustCommand\onedot{\futurelet\@let@token\@onedot}
\def\@onedot{\ifx\@let@token.\else.\null\fi\xspace}

\def\eg{e.g\onedot}

\makeatother

\definecolor{MyDarkBlue}{rgb}{0,0.08,0.8}
\definecolor{MyDarkGreen}{RGB}{45,155,45}
\definecolor{MyDarkRed}{rgb}{0.8,0.02,0.02}
\definecolor{MyOrange}{rgb}{1.0, 0.4, 0.2}
\definecolor{MyPurple}{RGB}{111,0,255}
\definecolor{MyRed}{rgb}{0.8,0.0,0.0}
\definecolor{MyGold}{rgb}{0.75,0.6,0.12}
\definecolor{MyDarkgray}{rgb}{0.66, 0.66, 0.66}

\newcommand{\ourstxt}{VDLM-txt\xspace} 
\newcommand{\oursmm}{VDLM-mm\xspace}
\newcommand{\ours}{VDLM\xspace} 
\newcommand{\oursfull}{VDLM\xspace} 
\newcommand{\bl}{Primal Visual Description\xspace} 
\newcommand{\blabbr}{PVD\xspace}

\newcommand{\imgtype}{\text{vector graphics}\xspace}

\newcommand{\ptloa}{Line or Angle\xspace} 
\newcommand{\ptaoo}{Angle Classification\xspace} 
\newcommand{\ptlc}{Length Comparison\xspace} 
\newcommand{\ptcq}{Clevr QA\xspace} 
\newcommand{\ptsw}{Shapeworld Scene\xspace} 
\newcommand{\ptmaze}{Maze Scene\xspace}

\newcommand{\dtaoo}{Angle Classification\xspace}
\newcommand{\dtlc}{Length Comparison\xspace} 
\newcommand{\dtsws}{Shapeworld Spatial Reasoning\xspace}
\newcommand{\dtswtobj}{Shapeworld Spatial Reasoning (2Obj)\xspace}
\newcommand{\dtswmobj}{Shapeworld Spatial Reasoning (MultiObj)\xspace} 
\newcommand{\dtswsup}{Shapeworld Superlative\xspace} 
\newcommand{\dtnlvr}{NLVR\xspace} 
\newcommand{\dtgeo}{Geoclidean 2-shot Learning\xspace} 
\newcommand{\dtmaze}{Maze Solving\xspace} 
\newcommand{\dtmazetwo}{2$\times$2 Maze Solving\xspace} 
\newcommand{\dtmazethree}{3$\times$3 Maze Solving\xspace} 
\newcommand{\dtvgbench}{VGBench-QA\xspace} 
\newcommand{\dtvgbenchcat}{VGBench-QA Category\xspace} 
\newcommand{\dtvgbenchcolor}{VGBench-QA Color\xspace} 
\newcommand{\dtvgbenchusage}{VGBench-QA Usage\xspace}

\definecolor{bggray}{HTML}{F5F5F5}
\definecolor{pvdblue}{HTML}{DAE8FC}
\usepackage{colortbl}
\usepackage{amsmath}

\definecolor{RoseQuartzBg}{HTML}{F7CAC9}
\definecolor{RoseQuartz}{HTML}{F5A798}
\definecolor{Serenity}{HTML}{92A8D1}
\definecolor{OrangeRed}{rgb}{1.0, 0.27, 0.0}
\definecolor{Turquoise}{HTML}{0F4C81}
\definecolor{mint}{rgb}{0.24, 0.71, 0.54}
\definecolor{byzantine}{rgb}{0.74, 0.2, 0.64}
\definecolor{byzantium}{rgb}{0.44, 0.16, 0.39}
\definecolor{captioningtask}{HTML}{9C843F}
\definecolor{qatask}{HTML}{CC6600}
\definecolor{temporalmarker}{HTML}{7F00FF}
\definecolor{targettext}{HTML}{3333FF}
\definecolor{prompttext}{HTML}{666666}
\definecolor{videolevel}{HTML}{330066}
\definecolor{framelevel}{HTML}{0066CC}
\definecolor{tokenlevel}{HTML}{336600}
\definecolor{boxgrey}{HTML}{666666}
\definecolor{boxblue}{HTML}{6C8EBF}
\definecolor{boxgreen}{HTML}{82B366}
\definecolor{textgreen}{HTML}{009900}
\definecolor{textred}{HTML}{FF0000}
\definecolor{textreddark}{HTML}{CC0000}
\definecolor{textblue}{HTML}{0066CC}

\usepackage{wrapfig}
\usepackage{xparse}
\usepackage{graphicx}
\usepackage{subcaption}

\usepackage{hyperref}
\usepackage{url}

\title{Visually Descriptive Language Model for Vector Graphics Reasoning\xspace}

\author{%
Zhenhailong Wang\textsuperscript{\textnormal{1}},
Joy Hsu\textsuperscript{\textnormal{2}}, 
Xingyao Wang\textsuperscript{\textnormal{1}}, 
Kuan-Hao Huang\textsuperscript{\textnormal{1,3}},
Manling Li\textsuperscript{\textnormal{2,4}}, \\
Jiajun Wu\textsuperscript{\textnormal{2}}, 
Heng Ji\textsuperscript{\textnormal{1}}
\\ 
\textsuperscript{1}University of Illinois Urbana-Champaign, \textsuperscript{2}Stanford University, \textsuperscript{3}Texas A\&M University, \textsuperscript{4}Northwestern University\\
{\small \{wangz3, xingyao6, khhuang, hengji\}@illinois.edu}\\
{\small \{joycj, manlingl\}@stanford.edu, jiajunwu@cs.stanford.edu}
}

\def\month{05}  
\def\year{2025} 
\def\openreview{\url{https://openreview.net/forum?id=WzS33L1iPC}} 

\begin{document}

\maketitle
\begin{center}
    \centering
    \captionsetup{type=figure}
    \includegraphics[width=0.98\textwidth]{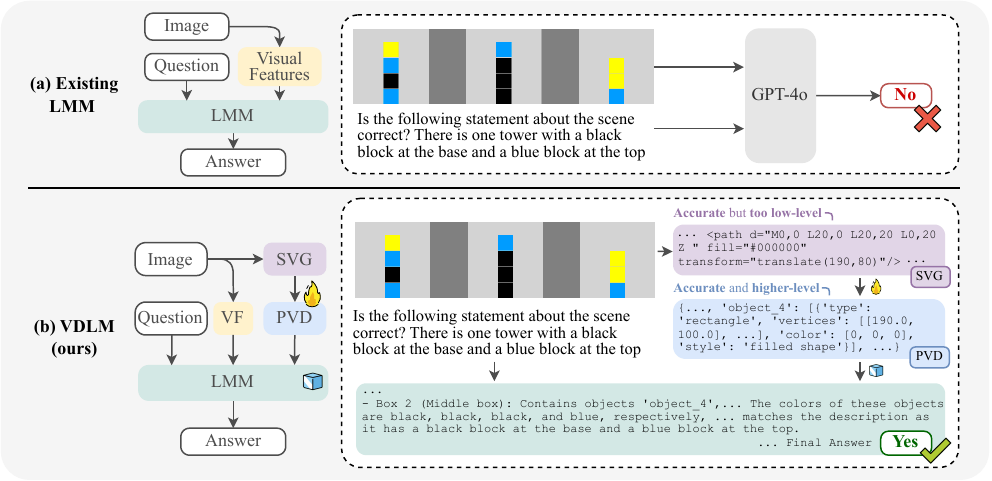}
    \caption{Existing monolithic LMMs rely solely on pretrained vision encoders, such as CLIP~\citep{clip}, for perception, which often fail to accurately capture low-level visual details in \imgtype-style images. In contrast, \ours{} enables precise visual reasoning by introducing SVG encoding and a learned intermediate symbolic representation, \bl (\blabbr{}), that bridges low-level SVG perception with high-level language reasoning. ``VF'' in (b) refers to ``visual features''.}
    \label{fig:teaser}
    \vspace{0.3cm}
\end{center}%

\begin{abstract}

\noindent Despite significant advancements, current large multimodal models (LMMs) struggle to bridge the gap between low-level visual perception---focusing on shapes, sizes, and layouts---and high-level language reasoning involving semantics, events, and logic. 
This limitation becomes evident in tasks requiring precise visual perception, such as comparing geometric properties or solving visual algorithmic reasoning problems.
To study this failure mode, we focus on an important visual domain: \imgtype---images composed purely of 2D objects and shapes, which are prevalent in web and mobile environments. 
Importantly, we consider \textit{rasterized} vector graphics without assuming access to their underlying vector code. 
We identify two key research questions: how can we enable precise visual perception, and how can we facilitate high-level reasoning based on such low-level perceptions?
To accurately capture low-level visual details, we explore using SVG for the precise encoding of visual scenes. 
However, SVGs are not readily interpretable by LLMs or LMMs in a zero-shot manner. 
To address this challenge, we propose the \textbf{Visually Descriptive Language Model (VDLM)} to build a bridge between low-level visual perception and high-level language reasoning.
VDLM learns an intermediate symbolic representation called \textbf{\bl (\blabbr)}, which translates raw SVGs into a higher-level abstraction comprising primitive attributes.
This abstraction allows for \textit{direct interpretation} by foundation models for zero-shot generalization to different reasoning tasks.
As an initial step to construct a descriptive intermediate representation for low-level visual reasoning, the SVG-to-\blabbr{} model is currently limited to simple compositions of primitive shapes, for which synthetic data can be generated without human annotation.
Nevertheless, empirical experiments show that \oursfull leads to significant improvements in state-of-the-art LMMs, such as GPT-4o, across various low-level multimodal perception and reasoning tasks on rasterized \imgtype.
Additionally, we provide extensive analyses of VDLM’s performance, showing that our framework offers improved interpretability due to its disentangled perception and reasoning processes. 
We also conduct an in-depth error analysis, highlighting remaining limitations and suggesting directions for future research.
Project page: \url{https://mikewangwzhl.github.io/VDLM/}

\end{abstract}

\section{Introduction}
In recent years, large multimodal models (LMMs)~\citep{openai2023gpt4v, team2023gemini, llava, chen2023internvl, bai2023qwen} have achieved impressive performance across a wide spectrum of general vision-language benchmarks~\citep{vqav2, fu2023mme, liu2023mmbench, yu2023mmvet, seed_bench}. However, these monolithic LMMs still struggle with seemingly simple tasks that require precise perception of low-level visual details.
In particular, we empirically observe that LMMs frequently exhibit this failure mode in \imgtype, which are images composed purely of 2D objects and shapes.
For example, a state-of-the-art LMM like GPT-4o~\citep{openai2024gpt4o} can still fail 43\% of the time when comparing the lengths of two line segments, and 54\% of the time when solving a simple 2$\times$2 maze.
LMMs’ ability to understand \imgtype is largely underexplored compared to natural images but is essential for growing downstream applications in web, visual design, and OS environments~\citep{zhou2023webarena, liu2024visualagentbench, xie2024osworld, rawles2024androidworld, zheng2024seeact, lu2024weblinx}.
In this work, we focus on the fundamental aspects of low-level visual reasoning involving \imgtype---including measurements, spatial relations, counting, and logical reasoning.
It is important to note that in this paper, ``vector graphics'' refers to \textbf{rasterized images} in JPEG or PNG format, without assuming access to their underlying vector code. This reflects a more realistic setting for visual reasoning in real-world scenarios, such as web~\cite{zhou2023webarena, lu2024weblinx} and mobile~\cite{wang2024mobile2, wang2025mobile}.
To address the aforementioned challenge, we identify two main research questions.
First, how can we enable precise visual perception in LMMs?
Second, how can we bridge the gap between low-level perception and high-level reasoning?

For our initial question, we explore encoding a rasterized image via vectorization with SVG representation, which describes a scene with paths (e.g., polygons and splines) and their corresponding measurements and positions. SVG representations, by nature, are unbiased towards high-level semantics and can capture low-level visual details in text.
The vectorization process can be faithfully accomplished with an rule-based raster-to-vector algorithm. 
However, such machine-generated SVG is often noisy and far from natural language, making it insufficient for language reasoning. Our preliminary experiments (\S\ref{sec_app:preliminary_exp}) demonstrate that existing foundation models are unable to interpret machine-generated SVG codes in zero-shot settings.
Another key challenge lies in the scarcity of end-to-end instruction tuning data containing $\langle$SVG, question, answer$\rangle$ triplets, making direct fine-tuning infeasible.

To bridge this perception-reasoning gap and address data scarcity, we propose translating the low-level SVG paths to a higher-level intermediate symbolic representation, referred to as \textbf{\bl(\blabbr)}, which can directly be leveraged by foundation models for multimodal reasoning.
Specifically, we learn an LLM-based~\citep{jiang2023mistral} SVG-to-\blabbr model, which transforms the raw SVG paths into a set of primitive attributes (e.g., shape, position) with corresponding predicted values (e.g., rectangle, pixel coordinates of the vertices). See Figure~\ref{fig:teaser} in the \colorbox{pvdblue}{blue} box for an example.
Notably, the \blabbr representation consists of primitive attributes that serve as fundamental building blocks for \imgtype, enabling learning from procedurally synthesized $\langle$SVG, \blabbr{}$\rangle$ pairs without requiring task-specific annotations. 
\blabbr{} filters out unnecessary noise in SVGs, enhancing the semantics of perception and facilitating subsequent reasoning.
Currently, we limit the scope of the SVG-to-PVD model to simple compositions of geometric primitives, for which synthetic data can be generated without human annotation. We leave generalization to entirely open-domain images as future work.

Comprising SVG-based image perception and \blabbr abstractions, we present the \textbf{Visually Descriptive Language Model (\oursfull)}, which contains three components: a rule-based visual encoder that converts images to SVG to capture precise visual details, a learned language model that translates SVG to \blabbr, and an inference-only reasoner that conducts zero-shot reasoning about downstream tasks with the \blabbr representation.
An overview of \ours is provided in Figure~\ref{fig:teaser}.
\begin{figure*}[t]
  \centering
  \includegraphics[width=0.99\textwidth]{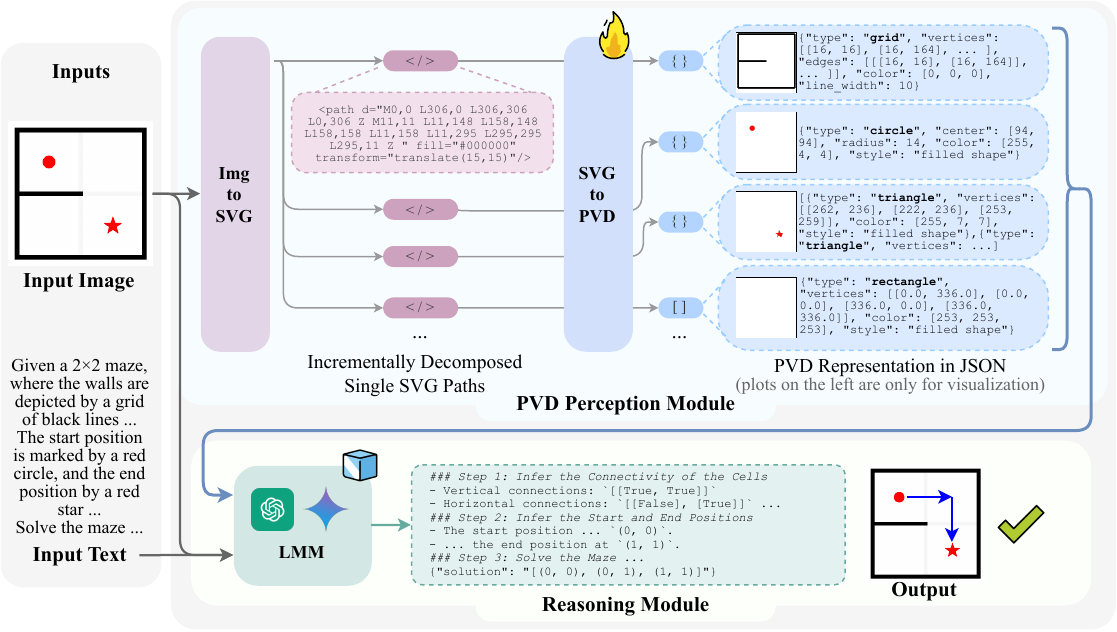}
  \caption{An example of \oursfull{} during inference. First, \oursfull{} extracts individual SVG paths from the input image and then transforms them into \bl (\blabbr{}) using a trained language model. These \blabbr{} perception results, along with the input text queries and the original input image, are subsequently fed into an LMM for reasoning.
  It is worth noting that although a “star” (\raisebox{-0.17em}{\includegraphics[height=1em]{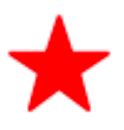}}) is not explicitly part of the \blabbr{} primitive ontology (see Figure~\ref{fig:base_layer}), the SVG-to-PVD model can approximate the “star” by composing two triangles (\raisebox{-0.17em}{\includegraphics[height=1em]{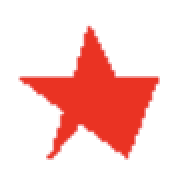}}). A strong off-the-shelf reasoner, such as GPT-4~\citep{openai2023gpt4}, can accurately deduce that this composition corresponds to the “star,” which is the target end position of the maze. For the complete response, refer to Figure~\ref{fig:full_response_of_maze_example}.
}
\vspace{-20pt}
\label{fig:overview}
\end{figure*}
Experimental results demonstrate that \ours achieves strong zero-shot performance in various visual reasoning tasks, outperforming LLaVA-v1.5~\citep{llava15}, G-LLaVA~\citep{gao2023gllava}, GPT-4V~\citep{openai2023gpt4v}, GPT-4o~\citep{openai2024gpt4o}, and Visual Programming approaches such as ViperGPT~\citep{suris2023vipergpt}.
Moreover, \ours also enhances interpretability through better disentanglement of perception and reasoning processes.

To summarize, the key contributions of our work are threefold:
First, we identify a critical failure mode of LMMs when reasoning about tasks that require precise, low-level perception.
Second, we introduce \ours, the first attempt to enhance LMMs' fine-grained visual reasoning capabilities with vectorized and symbolic visual descriptions---SVG representations and learned \bl.
Finally, we present an in-depth analysis of the emergent patterns, the disentangled impact of PVD quality on end-task performance, and the remaining limitations, highlighting directions for future work.

\section{\oursfull Framework}
\label{sec:method}

We introduce the \ours framework, which consists of three main components. First, a rule-based perception module converts images into SVG format, capturing low-level visual details (\S~\ref{subsec:image-to-svg}) that are highly complementary to CLIP-like image features. 
Second, a trained language model aligns SVGs with intermediate visual descriptions (\S~\ref{subsec:svg-to-baselayer}). Third, an inference-only foundational model reasons about downstream tasks using both visual and textual perception results (\S~\ref{subsec:baselayer-to-answer}). Refer to Figure~\ref{fig:overview} for an overview of \ours.

\subsection{Precise Visual Perception with SVG Encoding}
\label{subsec:image-to-svg}
Prior work~\citep{clip_fail_fine_retrieval, eyes_wide_shut} has demonstrated that, although CLIP-based~\citep{clip} vision encoders are effective at capturing high-level visual semantics, they can fall short in preserving fine-grained visual details.
We propose extracting an SVG representation that more accurately captures detailed measurements and is highly complementary to the widely used visual features. This can be achieved using rule-based image-to-SVG converters, such as Vtracer~\cite{VTracer}, for which we empirically observe near-perfect reconstruction quality on rasterized vector graphic images (see detailed analysis in Appendix~\ref{app:tmlr_added_vtracer_encoding_quality}).
SVG describes shapes, lines, and colors using mathematical expressions and paths with precise coordinates. 

We conduct a suite of preliminary experiments (\S~\ref{sec_app:preliminary_exp}) to investigate the potential of using SVG for representing visual inputs.
We empirically observe that SVG representation outperforms CLIP-based features on low-level \imgtype reasoning tasks given sufficient task-specific fine-tuning data.
However, two key challenges remain (\S~\ref{subsec_app:remaining_challenges_of_raw_svg_repre}):
First, off-the-shelf foundation models, such as GPT-4~\citep{openai2023gpt4}, have limited zero-shot reasoning abilities when dealing with raw SVG code.
Second, obtaining task-specific $\langle$SVG, question, answer$\rangle$ data for fine-tuning is extremely challenging, which restricts generalization to unseen tasks and domains.
We discuss how we address these challenges by learning an intermediate abstraction.

\begin{figure}[t]
  \centering
  \includegraphics[width=0.7\textwidth]{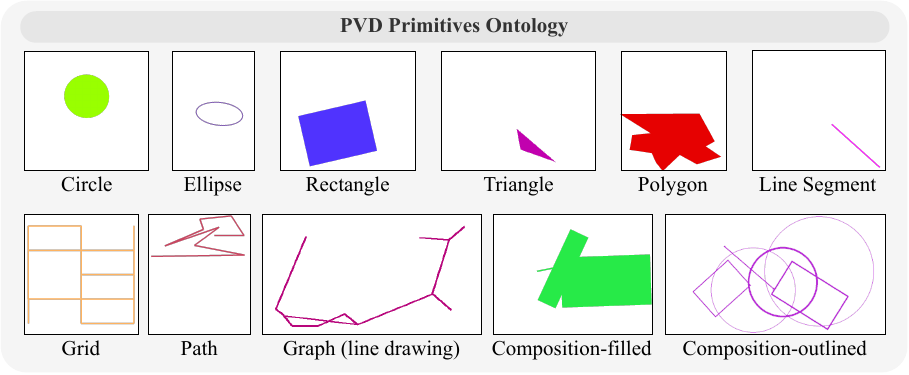}
  \caption{Ontology of the primitives in \blabbr{}. Composition of fundamental building blocks for \imgtype, as shown in Figure~\ref{fig:downstream_tasks}. }
  \vspace{-10pt}
  \label{fig:base_layer}
\end{figure}

\subsection{Bridging Low-Level Visual Perception with High-Level Language Reasoning}
\label{subsec:svg-to-baselayer}

\paragraph{\bl (\blabbr).}

We propose \bl, a higher-level abstraction that transforms low-level SVG paths to more structured primitives required for reasoning. \blabbr is a text-based visual description that consists of a set of primitive geometry objects, e.g., circles and line segments. Each \blabbr element contains the primitives' attributes (e.g., color, shape, position, size) with corresponding predicted values (e.g., blue, circle, pixel coordinates of the center, length of the radius). An example of the \blabbr representation is as follows (See Figure~\ref{fig:pvd_json_schema} for full definitions):

\begin{center}
\begin{tcolorbox}[colback=bggray, 
                  boxrule=0pt, 
                  arc=1.5mm,
                  width=0.48\textwidth, 
                  center title,
                  left=2mm, 
                  right=1mm, 
                  top=1mm, 
                  bottom=1mm] 
\small
\texttt{\{"type": "circle", "center": [252, 315], "radius": 202, "color": [175, 155, 98], "style": "filled shape"\}.
}
\end{tcolorbox}
\end{center}

Unlike raw SVG code, \blabbr can be \textit{directly reasoned about} by strong off-the-shelf foundation models to generalize across various downstream tasks. 
Moreover, \blabbr is sufficient to serve as a unified visual description across different types of \imgtype, as complex concepts can be composed of multiple primitive shapes. For example, a ``cross'' can be composed of two ``rectangles.''
As shown in Figure~\ref{fig:base_layer}, the ontology of the \bl contains 9 canonical primitive shape types that can be composed to cover various \imgtype in the wild. The primitive shapes include circles, ellipses, rectangles, triangles, polygons, line segments, grids, paths, and graphs. A path in \blabbr{} is defined as a non-intersecting polyline. Graphs and grids are defined as a set of vertices connected by a set of edges. 
As an initial step to build a visually descriptive intermediate representation, we focus on demonstrating proof of concept; extension to a more comprehensive ontology will be left for future work.

\vspace{-10pt}
\paragraph{Learning SVG-to-PVD alignment with a language model.}
We then train a language model to generate \blabbr outputs from SVG inputs. The input is a single SVG path depicting a visual concept, and the output is the predicted one or more primitives in the defined \blabbr ontology.
During inference, given an arbitrary raster image, we first convert it into a raw SVG file, which may contain a large number of SVG paths, including noise and speckles. To denoise the raw SVG file and extract salient shapes, we propose an incremental decomposition algorithm. 
Specifically, we incrementally include SVG paths while checking the difference between the partially rendered image of currently chosen paths and the fully rendered image of the original raw SVG file. We compute the summation of the absolute pixel-by-pixel difference between the two images and set an empirical threshold. If the difference after adding a new path is below this threshold, i.e., if the added path does not bring much additional visual information to the scene, we will skip that path. 
For the ordering of the path selection, we follow the default ordering from \cite{VTracer} that heuristically places the paths with a larger area at the front. 
The paths that come afterward will be stacked on top of previous paths during rendering. 
Upon obtaining the decomposed single SVG paths, we first generate their \blabbr representation individually. We then aggregate the individual \blabbr predictions into a holistic perception of the entire image using this JSON template: \texttt{["object\_0": <\blabbr output for path 0>, "object\_1": <\blabbr output for path 1>, ...]}.

Importantly, since \blabbr is \textbf{task-agnostic}, the data for training the SVG-to-\blabbr model can be procedurally generated without human annotation. 
We develop a data generator leveraging PIL.ImageDraw
and \cite{VTracer}, which creates a large-scale $\langle$SVG, \blabbr{}$\rangle$ paired dataset containing randomly generated primitives.
In some real-world tasks, such as geometry problems, multiple primitive shapes with the same color can overlap. When converted to SVG, these shapes tend to be parsed into one merged SVG path. To enable the SVG-to-\blabbr model to learn to decode individual primitives from such compositional concepts, we additionally generate data instances with randomly overlapped shapes. The target \blabbr representation, in this context, is a list of primitive \blabbr JSON objects. We ensure that each generated image contains only one unicolor object, single or composed, so that the converted SVG contains a single SVG path. This facilitates a language model in effectively learning the alignment between SVG and \blabbr.

To improve the robustness to unseen inference images, we randomize the image sizes, the positions and rotations of the shapes, as well as the styles of the shapes (filled or outlined). We additionally use two data augmentation methods, Gaussian Blur and Pixel Noise, to add variance to the training SVG paths. 
Our final dataset contains 160K $\langle$SVG, \blabbr{}$\rangle$ pairs. More details can be found in Appendix~\ref{sec_app:base_layer_details}.

We fine-tune a pretrained Mistral-7b~\citep{jiang2023mistral}
model on the synthesized \blabbr 160K dataset to perform SVG-to-\blabbr generation. We conduct full-parameter fine-tuning for 3 epochs with a learning rate of 1e-5.
The training objective is a standard Language Modeling loss on the generated \blabbr tokens as follows:
\begin{align}
    \mathcal{L} = -\frac{1}{N} \sum_{i=1}^{N} \log P(\mathbf{d}_i | \mathbf{s}, \mathbf{d}_{0:i-1})
\end{align} 
where $\mathbf{s}$ and $\mathbf{d}$ refer to the input SVG tokens and the generated \blabbr tokens respectively.
We use the Megatron-LLM~\citep{epfmgtrn} library for efficient LLM fine-tuning and the entire training process can be done in 16 hours on 4 NVIDIA A100-40GB GPUs.

\subsection{Enhancing Low-Level Visual Reasoning with Primal Visual Descriptions}
\label{subsec:baselayer-to-answer}
\blabbr provides precise visual details that are highly complementary to the semantic-centric visual features from pretrained visual encoders, such as CLIP. Its textual nature further facilitates direct integration into an inference-only LLM or LMM reasoner.

We explore two variants of \ours: \textbf{\oursmm} and \textbf{\ourstxt}, based on the type of reasoner applied. By default, \ours utilizes a multimodal LMM as the reasoner, referred to as \oursmm, which takes both the original image and the \blabbr perception as input. To examine the efficacy of using \blabbr to represent visual information, we also consider \ourstxt, which uses a text-only LLM as the reasoner.
A detailed execution trace of the \oursfull functions is illustrated in Figure~\ref{fig:overview}. We observe that a strong reasoner, such as GPT-4~\citep{openai2023gpt4}, without any fine-tuning, can effectively perform various types of task-specific reasoning based on the \blabbr representation. 
This includes identifying higher-level concepts, computing measurements, examining spatial relations, and performing multi-step reasoning.
The reasoning procedure is also more explainable and transparent compared to the output of existing monolithic LMMs.

\begin{figure*}[t]
  \centering
  \includegraphics[width=0.95\textwidth]{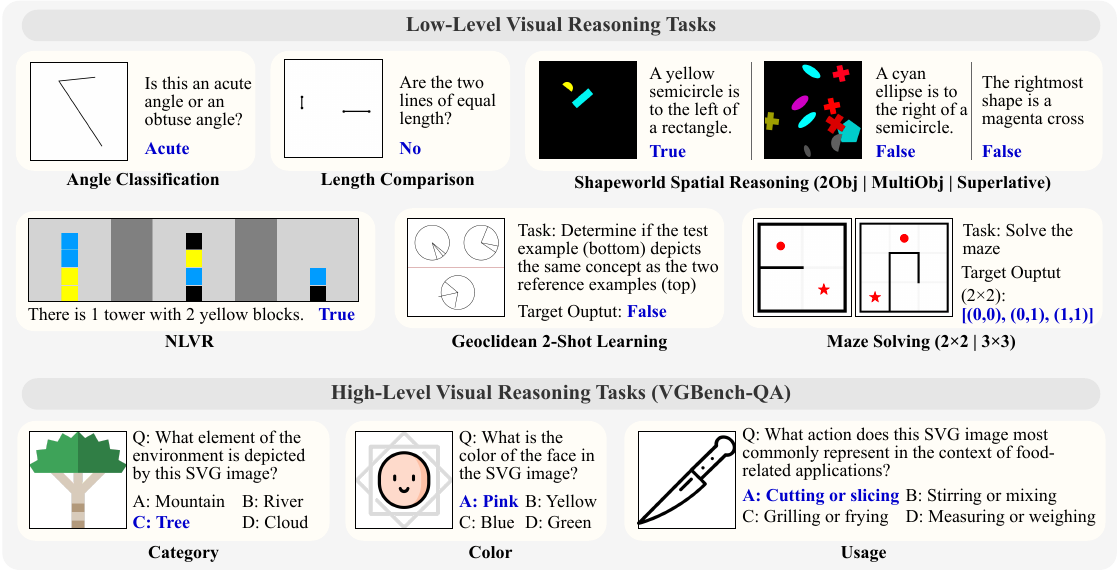}
  \caption{Our full evaluation benchmark with a focus on \textbf{low-level visual reasoning} about \imgtype (detailed in \ref{subsec:tasks}). We additionally include high-level reasoning tasks with rendered SVG images from VGBench-QA~\citep{zou2024vgbench}. All tasks are evaluated in a zero-shot setting.}
  \label{fig:downstream_tasks}
\end{figure*}

\vspace{-5pt}
\section{Experiments}
\label{sec:results}

\vspace{-2pt}

\subsection{Tasks}
\vspace{-2pt}
\label{subsec:tasks}
\paragraph{Low-level visual reasoning tasks.} Evaluating LMMs in tasks that require precise visual perception about \imgtype is a highly underexplored research area and has limited existing resources.
To this end, we construct a new evaluation benchmark that comprises 9 tasks which cover important aspects of low-level visual perception and reasoning, including measurements, spatial relations, counting, logical reasoning, and complex reasoning problems such as maze solving. 
The description of each task is as follows:
(1) \textbf{\dtaoo}: Identify whether an angle is acute or obtuse.
(2) \textbf{\dtlc}: Determine whether two line segments are of equal length.
(3-4) \textbf{\dtsws}: The Shapeworld~\citep{kuhnle2017shapeworld} dataset on spatial relations with images containing exactly two objects or multiple objects.
(5) \textbf{\dtswsup}: The Shapeworld dataset on superlative statements.
(6) \textbf{\dtnlvr}: The Natural Language for Visual Reasoning dataset~\citep{NLVR} which contains diverse counting, spatial reasoning, and logical reasoning queries.
(7) \textbf{\dtgeo}: A repurposed Geoclidean~\citep{hsu2022geoclidean} dataset requiring the model to understand a compositional geometric concept with only two reference examples.
(8-9) \textbf{\dtmaze}: Solve a 2$\times$2 or 3$\times$3 maze, given the starting and ending positions.
Among these tasks, \dtaoo, \dtlc, and \dtmaze are newly created from scratch (See Appendix~\ref{sec_app:new_downstream_tasks} for more details).

\vspace{-16pt}
\paragraph{High-level visual reasoning tasks.} Although the focus of this work is on low-level visual reasoning, we additionally  include a set of high-level tasks to investigate the impact of \ours on knowledge reasoning tasks. These tasks rarely require precise perception of the locations and measurements of the primitives.
We leverage VGBench~\citep{zou2024vgbench}, a benchmark originally proposed for evaluating LLMs in understanding and generating vector graphics codes. In this work, we evaluate LMMs and \oursmm for question-answering based on the rasterized VGBench SVG images. 

Figure~\ref{fig:downstream_tasks} shows simplified input and output examples for each task. Full prompts can be found in Appendix~\ref{sec_app:full_prompts}.
To reduce the cost of evaluating proprietary models, we randomly sample a subset of 100 instances for each task.
We consider a zero-shot evaluation setting for all tasks. Note that the SVG-to-\blabbr{} model in \ours{} is trained purely on synthesized task-agnostic data.

\begin{table*}[t]
\centering
\resizebox{0.95\textwidth}{!}
{
\centering

\begin{tabular}{lc|ccccccccc|c}
    \toprule
    \multicolumn{12}{c}{\textbf{Low-level Visual Reasoning on Vector Graphics}}\\
    \toprule
    & \textbf{\begin{tabular}[c]{@{}c@{}} Tools \end{tabular}} &
        \textbf{\begin{tabular}[c]{@{}c@{}} AC \end{tabular}} &
        \textbf{\begin{tabular}[c]{@{}c@{}} LC \end{tabular}} &
        \textbf{\begin{tabular}[c]{@{}c@{}} SW-S\\2Obj \end{tabular}} &
        \textbf{\begin{tabular}[c]{@{}c@{}} SW-S\\mObj \end{tabular}}&
        \textbf{\begin{tabular}[c]{@{}c@{}} SW\\Sup \end{tabular}} &
        \textbf{\begin{tabular}[c]{@{}c@{}} NLVR \end{tabular}} &
        \textbf{\begin{tabular}[c]{@{}c@{}} Geo \end{tabular}} &
        \textbf{\begin{tabular}[c]{@{}c@{}} Maze\\2$\times$2 \end{tabular}} & 
        \textbf{\begin{tabular}[c]{@{}c@{}} Maze\\3$\times$3 \end{tabular}} & 
        \textbf{\begin{tabular}[c]{@{}c@{}} All \end{tabular}} \\
    \midrule
    \rowcolor{gray!20}
    \multicolumn{12}{c}{Monolithic Large Multimodal Models}\\
    Llava-1.5-7b & - & 0.53 & 0.49 & 0.48 & 0.55 & 0.35 & 0.53 & 0.50 & 0.00 & 0.00 & 0.381 \\
    Llava-1.5-13b & - & 0.53 & 0.51 & 0.51 & 0.47 & 0.61 & 0.48 & 0.50 & 0.00 & 0.00 & 0.401 \\
    Gllava-7b & - & 0.59& 0.50& 0.43& 0.54& 0.43& 0.49& 0.58& 0.00& 0.00& 0.396 \\
    \rowcolor{blue!10} GPT-4V & - & 0.58 & 0.64 & 0.77 & 0.60 & 0.61 & 0.63 & 0.64 & 0.28 & 0.02 & 0.530\\
    \rowcolor{orange!10} GPT-4o & - & 0.63& 0.57& \textbf{0.97}& \textbf{0.82}& \textbf{0.92}& 0.81& \textbf{0.71}& 0.46& 0.08& 0.663\\
    \midrule
    \rowcolor{gray!20} \multicolumn{12}{c}{Visual Programming with LLM (text-only) reasoner}\\
    ViperGPT (w/ GPT-4) & CI & 0.11 & 0.67 & 0.61 & 0.47 & 0.53 & 0.43 & 0.02 & 0.03 & 0.00 & 0.319\\
    \midrule
    \rowcolor{gray!20} \multicolumn{12}{c}{\ours with LLM (text-only) reasoners}\\
    \textbf{\ourstxt} (w/ GPT-4) & - & 0.89 & 0.95 & 0.78 & 0.63 & 0.80 & 0.68 & 0.63 & 0.40 & 0.19 & 0.661\\
    \textbf{\ourstxt} (w/ GPT-4) & CI & 0.73 & 0.95 & 0.89 & 0.68& 0.72 & 0.72 & 0.64 & 0.40 & 0.26 & 0.666\\
    \midrule
    \rowcolor{gray!20}\multicolumn{12}{c}{\ours{} with LMM (multimodal) reasoners}\\
    \rowcolor{blue!10} \textbf{\oursmm} (w/ GPT-4V) & - & 0.55& 0.94& 0.84& 0.62& 0.72& 0.71& 0.69& 0.60& 0.20& 0.652\\
    \rowcolor{orange!10} \textbf{\oursmm} (w/ GPT-4o) & - & \textbf{0.90} & \textbf{0.95} & 0.91 & \textbf{0.82} & 0.82& \textbf{0.86} & \textbf{0.71} & \textbf{0.61} & \textbf{0.34} & \textbf{0.769}\\
    \bottomrule
\end{tabular}
}
\caption{Zero-shot accuracy on low-level visual reasoning tasks. Task abbreviations: AC (\dtaoo), LC (\dtlc), SW-S-2Obj/mObj (\dtsws with two objects or multiple objects), SW-Sup (\dtswsup), Geo (\dtgeo). ``CI'' refers to Code Interpreter.
\oursmm brings consistent overall improvements to GPT-4V and GPT-4o, as indicated by the comparison within the \colorbox{blue!10}{blue} and \colorbox{orange!10}{orange} rows.
Detailed analysis is presented in \S\ref{subsec:results} and \S\ref{sec:analysis}.
}
\vspace{-10pt}
\label{tab:main_result}
\end{table*}

\begin{table}[t]
\centering
\resizebox{0.56\textwidth}{!}
{
\centering
\begin{tabular}{@{}lcccc}
    \toprule
    \multicolumn{5}{c}{\textbf{High-level Visual Reasoning on Vector Graphics}}\\
    \toprule
    & \multicolumn{4}{c}{\textbf{VGBench-QA}} \\
    & \textbf{\begin{tabular}[c]{@{}c@{}} Category \end{tabular}} &
        \textbf{\begin{tabular}[c]{@{}c@{}} Color \end{tabular}} &
        \textbf{\begin{tabular}[c]{@{}c@{}} Usage \end{tabular}} &
        \textbf{\begin{tabular}[c]{@{}c@{}} All \end{tabular}}\\
    \midrule
    Llava-v1.5-7b & 0.26& 0.32& 0.27& 0.283 \\
    Llava-v1.5-13b & 0.32& 0.43& 0.39& 0.380 \\
    Gllava-7b & 0.16 & 0.33& 0.21& 0.233 \\
    GPT-4o & 0.58 & 0.84 & \textbf{0.76}& 0.726\\
    \textbf{\oursmm} (w/ GPT-4o) & \textbf{0.62} & \textbf{0.86} & 0.75& \textbf{0.743}\\
    \bottomrule
\end{tabular}
}
\caption{Zero-shot accuracy on high-level visual reasoning tasks. We show that \oursmm preserves the LMM's capability on semantic-centric reasoning that does not require precise low-level perception.}
\vspace{-10pt}
\label{tab:vgbench_results}
\end{table}

\vspace{-5pt}
\subsection{Models}
\vspace{-3pt}
We compare our work with strong baselines, including both state-of-the-art monolithic large multimodal models (LMMs), i.e., LLaVA-v1.5~\citep{llava15}, GLLaVA~\cite{gao2023gllava}, GPT-4V~\citep{openai2023gpt4}\footnote{GPT-4V model version: gpt-4-1106-vision-preview.}, GPT-4o~\citep{openai2024gpt4o}\footnote{GPT-4o model version: gpt-4o-2024-05-13}, as well as visual programming agents, e.g., ViperGPT~\citep{suris2023vipergpt}.
ViperGPT employs an LLM to generate code, which can call external vision models, such as GLIP~\citep{glip} and BLIP2~\citep{blip2}, to process the image and generate the final output. 
Given that ViperGPT-style models successfully separate perception from reasoning, we seek to investigate whether the existing perception tools adequately recognize low-level primitives in \imgtype.
For \oursfull, we explore two variants, namely \oursmm with GPT-4V, GPT-4o and \ourstxt with GPT-4 (text-only).
\footnote{GPT-4 (text-only) model version: gpt-4-0125-preview.}
We also experiment with applying weaker LMM reasoners, such as LLaVA, to \oursmm{}. We find that interpreting \blabbr{} requires a certain level of text reasoning capability, and the benefits only emerge with strong LMMs, as shown in Figure~\ref{fig:lmm_improv}. However, recent strong open-source LMMs, such as Qwen-2.5-VL-72B~\citep{qwen25vl}, can also benefit from PVD features (see Appendix~\ref{app:tmlr_added_open_source_lmm_reasoner} for additional experiments).
To obtain more insights in comparing with ViperGPT, we further investigate augmenting \ourstxt with a Code Interpreter (CI). 
We employ the GPT-4 Assistant~\footnote{\url{https://platform.openai.com/docs/assistants/overview/agents}} for our experiments, designating the code interpreter as the sole tool available.
We use the same set of prompts for both \ourstxt and \oursmm.
See details about prompt design in Appendix~\ref{sec_app:full_prompts}.

\vspace{-2pt}
\subsection{Results} 
\label{subsec:results}
\vspace{-3pt}

Table~\ref{tab:main_result} shows the zero-shot accuracy for the evaluation tasks. We outline the key findings as follows: 


\begin{wrapfigure}{r}{0.5\textwidth}
  \centering
  \vspace{-15pt}
  \includegraphics[width=0.48\textwidth]{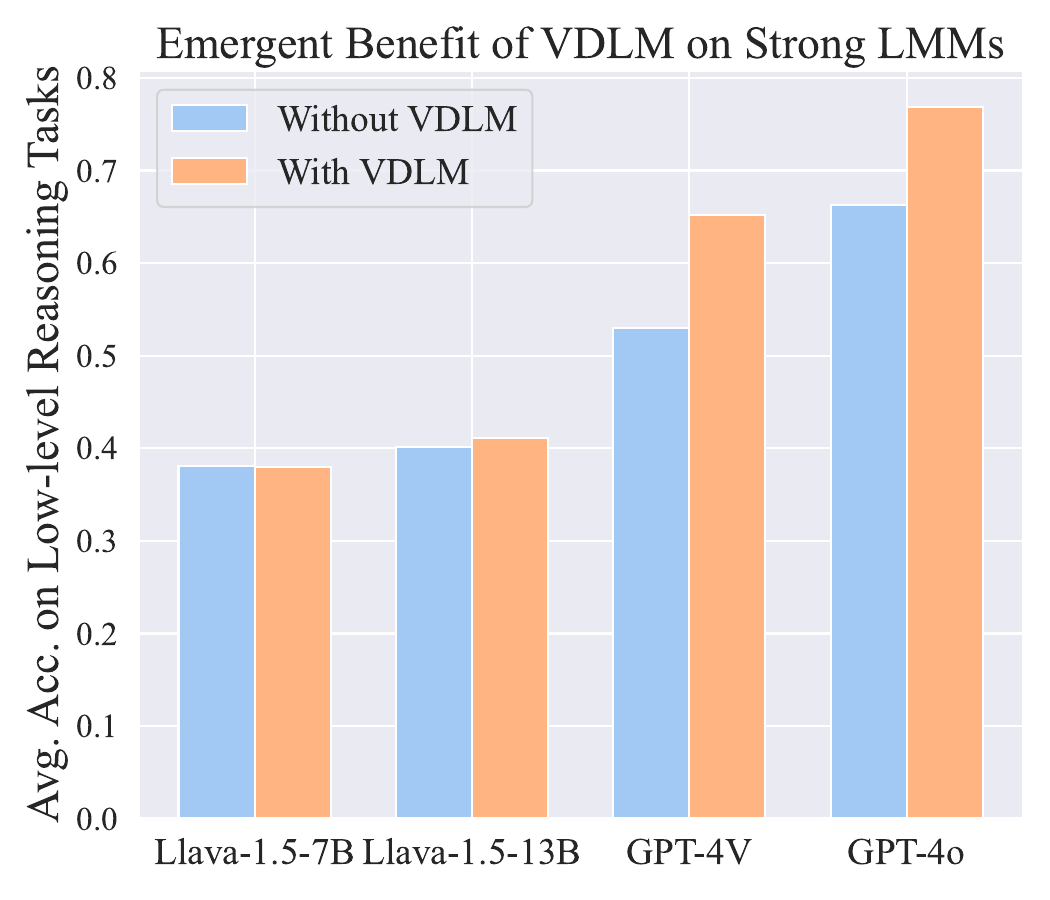}
  \caption{The direct improvements brought by \ours to LMMs emerge when the LMM possesses sufficient text reasoning capabilities. These improvements are consistent with stronger LMMs, such as GPT-4o, which have enhanced spatial reasoning performance.}
  \vspace{-15pt}
  \label{fig:lmm_improv}
\end{wrapfigure}

\paragraph{\ours significantly improves LMMs on low-level reasoning tasks, while preserving their capabilities in high-level reasoning.} 
The fact that \ourstxt performs competitively even with a text-only reasoner highlights the efficacy of the intermediate \blabbr{} representation for precise low-level visual perception.
Without any task-specific fine-tuning, strong LMM reasoners can effectively incorporate the additional information provided \blabbr{} alongside the image input. 
Figure~\ref{fig:lmm_improv} further demonstrates that this benefit only emerges when the LMM has a certain level of text-reasoning ability and persists in state-of-the-art LMMs. For high-level reasoning tasks (Table~\ref{tab:vgbench_results}), the improvement is more subtle, as the tasks focus on the semantics of the \imgtype, such as “what can this be used for?”, which rarely require precise location or measurements of visual elements.

\paragraph{QA performance on complex math problems does not necessarily reflect a faithful understanding of low-level visual concepts.} 
We observe that G-LLaVA~\citep{gao2023gllava}, a model demonstrating strong performance on geometric problems, such as MathVista~\citep{lu2023mathvista}, still struggles with understanding basic lines and angles, which are prerequisites for solving geometric math problems. 

\vspace{-5pt}
\paragraph{Existing vision-language models, such as GLIP and BLIP2, are ineffective as low-level visual preceptors.} This is evidenced by the unsatisfactory performance of ViperGPT, even when equipped with a strong planner like GPT-4. On the other hand, we observe that augmenting the reasoning model in \ourstxt with code interpreters can be particularly helpful for tasks requiring algorithmic reasoning, such as 3$\times$3 maze solving.

\vspace{-5pt}
\paragraph{Impact of errors in PVD perception.} In certain tasks, such as Shapeworld Spatial Reasoning, GPT-4o achieves better performance than \oursmm{}. 
The reason lies in the imperfect perception results of the SVG-to-\blabbr{} model, despite the image-to-SVG step achieving near-perfect reconstruction as shown in Appendix~\ref{app:tmlr_added_vtracer_encoding_quality}.
Since the SVG-to-\blabbr{} model is trained with purely synthetic data, it is not yet perfect when generalizing to diverse domains. 
In Appendix~\ref{app:tmlr_added_llm_choice_and_png_to_pvd}, we present an additional ablation study on the design choices of the PVD model, including alternative LLM selections and a comparison to the PNG-to-PVD approach.
We also carefully analyze the remaining errors in \S\ref{subsec:analysis_errors}, and demonstrate the impact of improving perception on end-task performance (\S\ref{subsec:analysis_perception_impact}). In \S\ref{sec:future_work}, we discuss future work for developing a more general and expressive \blabbr{} representation.

\vspace{-5pt}
\section{Analysis}
\label{sec:analysis}

\begin{figure}[t]
    \centering
    \begin{minipage}{0.5\textwidth}
        \centering
        \includegraphics[width=\textwidth]{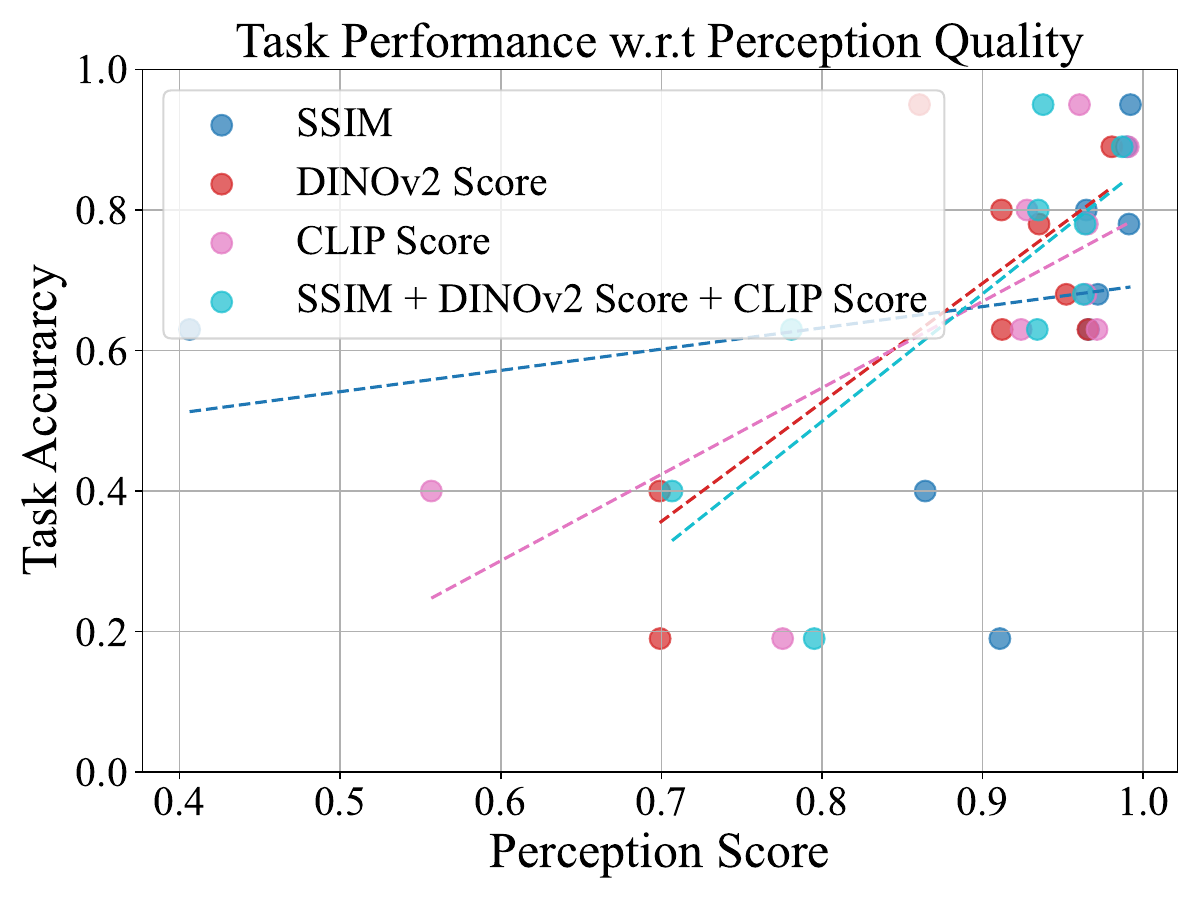}
        \caption{Task-level correlation between \blabbr{} perception quality and end-task performance.}
        \label{fig:task_level_perception_vs_performance}
    \end{minipage}
    \hfill
    \begin{minipage}{0.46\textwidth}
        \centering
        \includegraphics[width=\textwidth]{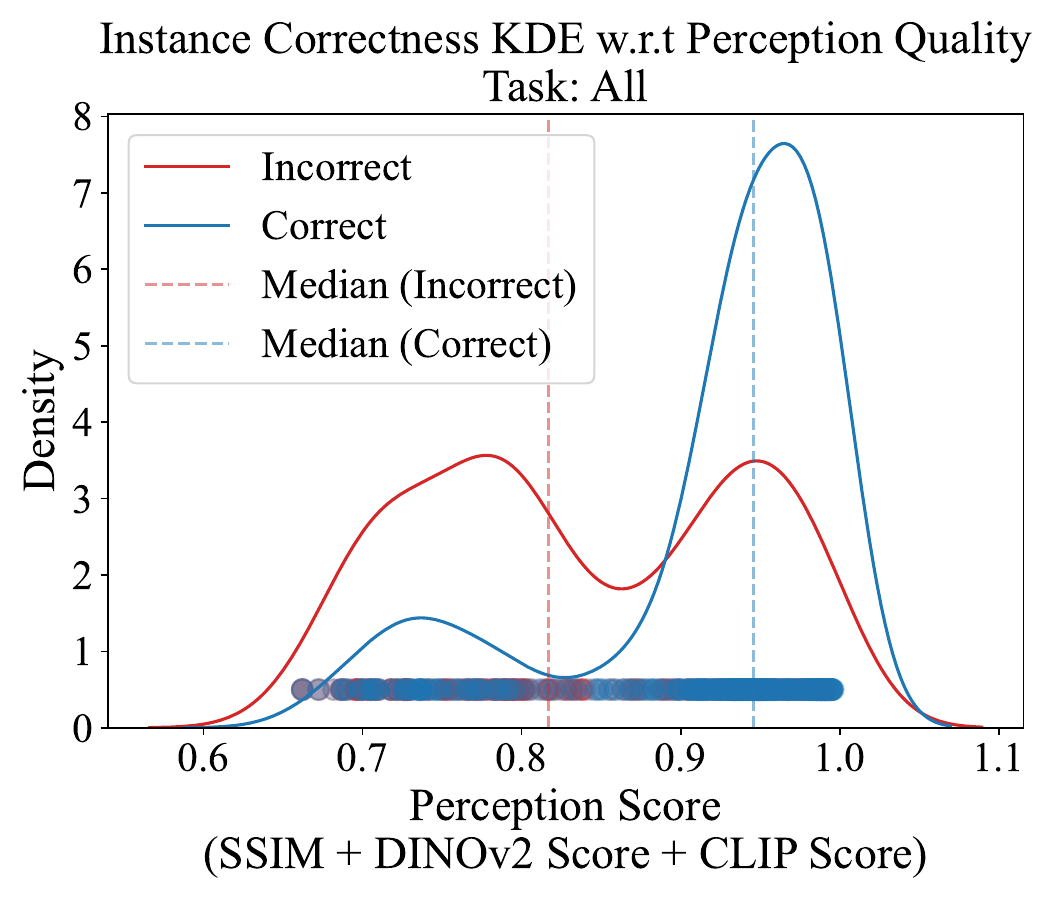}
        \caption{Instance-level correlation between \blabbr{} perception quality and end-task performance. We observe a consistent positive correlation as in task-level.}
        \label{fig:instance_level_perception_vs_performance}
    \end{minipage}
    \vspace{-10pt}
\end{figure}

\vspace{-2pt}

\subsection{\bl Quality vs End-Task Performance}
\label{subsec:analysis_perception_impact}
One advantage of a modular system is that enhancing an individual module can leads to improvements in the overall system. In this section, we explore whether a positive correlation exists between the quality of the intermediate perception representation and end-task performance. To investigate this, we first define metrics to reflect the quality of the \blabbr perception. Upon generating a \blabbr perception result, we render it back into a raster image using our procedural image generator. We then compute a similarity score between the reconstructed image and the original input image as a measure of the perception performance. For measuring the similarity, we consider both pixel-based and embedding-based metrics. We adopt the Structural Similarity (SSIM) Index~\citep{ssim} score to assess pixel-level similarity. Additionally, to account for semantic similarity, we adopt a CLIP-score~\citep{clip} and a DINOv2-score~\citep{oquab2023dinov2}, which are calculated as the cosine similarity of the flattened CLIP and DINOv2 embeddings, respectively.

In Figures~\ref{fig:task_level_perception_vs_performance} and \ref{fig:instance_level_perception_vs_performance},  
we visualize the impact of the perception quality, on the 9 low-level reasoning tasks with \ourstxt, at both the task and instance levels. In Figure~\ref{fig:task_level_perception_vs_performance}, each point denotes the accuracy of a task, with different colors representing different similarity metrics. 
The dashed lines depict linear regression results of the points, revealing a consistent positive correlation between perception quality and task accuracy across the metrics. Since the task-level accuracy may not be directly comparable across different tasks, we additionally perform an instance-level analysis using Kernel Density Estimation (KDE) on the correctness of all task instances with respect to their perception scores. 
As shown in Figure~\ref{fig:instance_level_perception_vs_performance}, the ``correct'' distribution visibly skews to the area of higher perception scores, indicating that better perception tends to result in a correct final answer. 
This finding is promising, suggesting that enhancing the intermediate \blabbr{} representation, even with a fixed reasoning model, can effectively boost downstream task performance.

\vspace{-3pt}
\subsection{Interpretable Error Analysis} 
\vspace{-1pt}
\label{subsec:analysis_errors}
The improved interpretability, resulting from \blabbr{}'s disentangled perception and reasoning, allows us to conduct an in-depth analysis of the failure modes of \oursfull. 
We find that both the perception step (SVG-to-\blabbr) and the reasoning step (\blabbr{}-to-answer) can contribute to errors. 
On tasks that require complex multistep reasoning, such as \dtmaze, reasoning errors become more prevalent; otherwise, perception errors most directly contribute to poor performance. 
Details and illustrative examples of these errors are provided in \textbf{Appendix~\ref{sec_app:error_analysis_details}}, along with a distribution of perception and reasoning errors from human analyses. The prevalent error types for both perception and reasoning steps are summarized as follows.

Common perception errors include failures in faithfully perceiving novel shapes that are not covered by or cannot be composed within the \blabbr ontology, and
failures in capturing intentional constraints between primitives, such as a line exactly segmenting a circle, due to the random nature of the data generation on the positioning of objects.
In Table~\ref{tab:impact_aug}, we show that the proposed augmentation during synthetic data generation improves \blabbr{} perception. 
Common reasoning errors over the \blabbr perception include failures in discovering intentional constraints without being explicitly asked, such as automatically recognizing that a rhombus is not the same concept as a general quadrilateral; failure in handling ambiguous instructions; and failure in complex multi-step reasoning tasks like solving mazes.

\section{Related Work}

\paragraph{Visual shortcomings in large multimodal models.}
While state-of-the-art LMMs achieve strong performance on existing multimodal benchmarks~\citep{vqav2, fu2023mme, llava, liu2023mmbench, yu2023mmvet, seed_bench}, which primarily focus on natural images, recent work~\citep{lu2023mathvista, yue2023mmmu, vlmchart, chart2023, hsu2022geoclidean, gao2023gllava} has shown that they struggle with charts, geometric diagrams, and abstract scenes.
This observation aligns with recent studies investigating visual shortcomings in LMMs. \cite{eyes_wide_shut} suggests that current LMMs struggle with visual details because the image-text contrastive pretraining of the CLIP visual backbone does not encourage the preservation of fine-grained visual features, such as orientation and quantity. 
To address this issue, recent studies have either leveraged the mixture-of-experts approach~\citep{eyes_wide_shut, fan2024mousi, lu2024deepseek, jain2023vcoder}, incorporating various types of vision encoders, such as SAM~\citep{sam}, DINOv2~\citep{oquab2023dinov2}, or introduced auxiliary losses that emphasize local details during multimodal pretraining~\cite{mckinzie2024mm1, SPARC, villa}.
In this work, we propose a novel perspective for addressing this visual deficiency in \imgtype with an intermediate perception representation.

\paragraph{Image vectorization and program synthesis.}
Generating vectorized or symbolic representations of visual concepts has been a topic of interest in both the NLP and computer vision communities. 
Recent work~\citep{vinker2022clipasso, sketch_representation, LIVE_vectorization, rodriguez2023starvector, jain2023vectorfusion,  tang2024strokenuwa, xing2024svgdreamer, hu2024vectorpainter} has investigated generating \imgtype codes from raster images or text prompts.
We focus on the reverse problem of understanding and reasoning about \imgtype as visual inputs. We find that \imgtype reasoning serves as a challenging testbed to evaluate low-level visual reasoning abilities in large multimodal models (LMMs).
Although \cite{gpt4_analysis_sparks, svgllm, zou2024vgbench, qiu2024can} have demonstrated the potential of text-only large language models (LLMs) in understanding the semantics of \imgtype codes, it remains unclear how to enhance the ability of large \textit{multimodal} models to process \textit{rasterized} vector graphics without access to the underlying code, which is more common in real-world scenarios.
Therefore, we propose the intermediate \bl{} representation to further enhance low-level perception and reasoning, without sacrificing the performance of semantic understanding.
This work is also heavily inspired by related work in neural-symbolic models~\citep{ritchie2016neurally_guided, wu2017neural_scene_derendering, yi2018neural_symbolic_vqa, mao2019neuro, hsu2023s, zhang2023editing_motion_graphics, alphageometry}.
This paradigm aims to de-render visual scenes into structured representations, retrieve programs from the input text, and execute these programs on the image representations.
However, most neural-symbolic work can not be applied to recent LMMs and is limited to specific tasks. We aim to learn a task-agnostic visual description that can be directly reasoned about by off-the-shelf LMMs.

\paragraph{Disentangling perception and reasoning in large multimodal models.}
Another closely related line of work has investigated disentangling visual perception and reasoning with visual programming~\citep{visprog, suris2023vipergpt, ge2023recursive_vp, vsearch} and tool-using~\citep{visual_chatgpt, liu2023llava_plus, qiao2024prism}.
These models leverage the code generation capabilities of LLMs to compose and employ a set of vision-language or vision-only models, such as object detection and image caption models, as subroutines for solving visual reasoning tasks. Despite promising performance on natural images, as shown in \S~\ref{sec:results}, we find that these models are still limited by the existing vision-language models' inability to process low-level primitives effectively.

\section{Limitations and Future Work}
\label{sec:future_work}

\ours represents an initial step to constructing a descriptive intermediate representation for low-level visual reasoning. While our results strongly support this proof of concept, several limitations remain, highlighting directions for future work.

First, due to the scarcity of end-to-end instructional tuning data (discussed in \S\ref{subsec:image-to-svg}), the current SVG-to-PVD model is constrained to compositions of simple primitives, for which we can synthesize data without requiring annotation. To enhance PVD's generalization across more diverse domains, we identify several possible improvements. For instance, incorporating human-created vector graphics with procedural annotations could enrich the ontology and training dataset. Additionally, integrating visual search~\cite{vsearch} during inference could facilitate the conversion of focused regions or parts into PVD, enabling multi-step perception and reasoning. Second, SVG and PVD are inherently designed for representing 2D vector graphics. Future research is needed to develop a more general intermediate representation that extends beyond 2D vector graphics to encompass 3D structures and natural images.

\section*{Acknowledgments}
This research is based upon work supported by U.S. DARPA ECOLE Program No. \#HR00112390060, AFOSR YIP FA9550-23-1-0127, ONR N00014-23-1-2355, ONR YIP N00014-24-1-2117, ONR MURI N00014-24-1-2748, and the Stanford Institute for Human-Centered AI (HAI). The views and conclusions contained herein are those of the authors and should not be interpreted as necessarily representing the official policies, either expressed or implied, of DARPA, or the U.S. Government. The U.S. Government is authorized to reproduce and distribute reprints for governmental purposes notwithstanding any copyright annotation therein.

\bibliography{main}
\bibliographystyle{tmlr}

\clearpage

\appendix

\section*{Appendix}

\appendix

The appendix is organized as follows: In Appendix~\ref{sec_app:preliminary_exp}, we present preliminary experiments comparing SVG and image-based representations. In Appendix~\ref{sec_app:error_analysis_details}, we include details on error analyses, and in Appendix~\ref{sec_app:base_layer_details}, we describe \bl details. 
Appendices~\ref{app:tmlr_added_vtracer_encoding_quality}–\ref{app:tmlr_added_open_source_lmm_reasoner} present additional experiments on Vtracer visual encoding quality, PVD model variants, and open-source LMM reasoner variants. Appendix~\ref{app:tmlr_added_pvd_parsing_novel_concept} provides additional qualitative examples of PVD parsing novel concepts via composition. Appendix~\ref{app:tmlr_added_prompt_engineering} presents further efforts in refining prompts for integrating PVD into LMMs.
Appendix~\ref{sec_app:full_response_of_overview_example} shows the full input and output from GPT-4 for the maze-solving example depicted in Figure~\ref{fig:overview}.
Task prompts and newly constructed downstream task datasets can be found in Appendices~\ref{sec_app:full_prompts} and~\ref{sec_app:new_downstream_tasks}, respectively. In Appendix~\ref{sec_app:dataset_stat}, we include detailed statistics for all of the datasets we used.

\section{Preliminary Experiments on SVG Representations}
\label{sec_app:preliminary_exp}

We introduce a suite of probing tasks to evaluate current LMMs' capabilities in performing tasks with \imgtype. The results show that even state-of-the-art LMMs, such as GPT-4V, struggle with tasks that require precise perception of low-level primitives, such as comparing the lengths of two lines. 
We then investigate where this deficiency originates and propose an alternative representation, Scalable Vector Graphics (SVG), for representing such precise low-level features. 
We find that, compared to image-based representations, SVG representations can be more efficient for visual reasoning on \imgtype. However, they are not without their own limitations, which we will elaborate on in \S~\ref{subsec_app:remaining_challenges_of_raw_svg_repre}.

\subsection{Image and SVG Representations}
\label{subsec_app:svg_probing_exp}

\begin{figure*}[t]
  \centering
  \includegraphics[width=\textwidth]{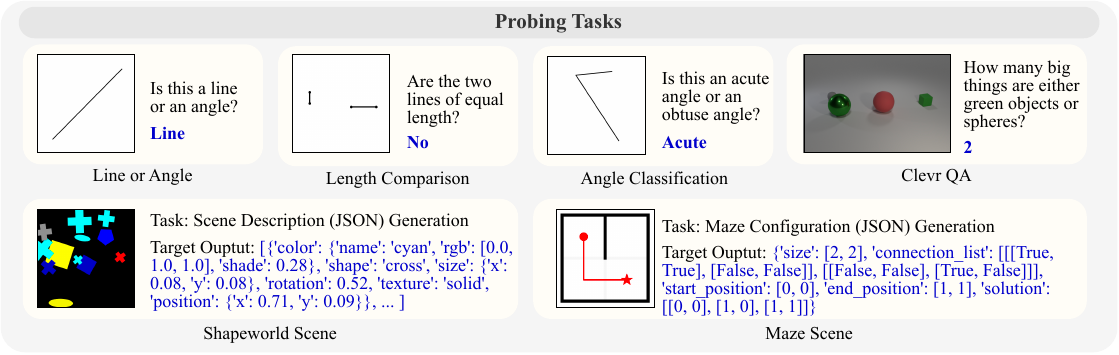}
  \caption{Illustration of the probing tasks. The four tasks at the top are question-answering tasks, while the two tasks at the bottom are scene-generation tasks. The goal of the scene-generation tasks is to generate the entire structured scene description following a predefined schema.}
  \label{fig:probing_tasks}
\end{figure*}

In the probing tasks, we include both discriminative and generative tasks, each with varying levels of emphasis on low-level visual details. Illustrations of the input and output examples are available in Figure~\ref{fig:probing_tasks}. 
We additionally include a non-vector-graphics task, \ptcq, which consists of realistic 3D rendered scenes. This is to test the limits of SVG representations in encoding 3D objects within realistic images.
Detailed statistics of these tasks can be found in Table~\ref{tab:dataset_stats}.

For each task, we consider two evaluation settings: zero-shot and fine-tuning. We explore two types of representations for the input image: (1) direct use of the image pixels, encoding them into patch embeddings with an image encoder, e.g., CLIP~\citep{clip}; (2) conversion of the image into SVG code using a rule-based raster-to-SVG converter~\citep{VTracer}.

For fine-tuning with the image input, we instruction-finetune Llava-v1.5-7b (including the LLM-backbone and the projection layer) using Lora~\citep{hu2021lora} on the training set for one epoch. For fine-tuning with the SVG input, we only fine-tune the LLM backbone of Llava-v1.5, Vicuna~\citep{vicuna2023}, using Lora for one epoch, with the input image's SVG code concatenated in the context. The results are shown in Table~\ref{tab:probing_results}. Key observations include:

(1) The SOTA open-source LMM, Llava-v1.5, struggles to achieve non-trivial performance on most probing tasks even with dedicated fine-tuning. On tasks with binary choices, Llava tends to predict homogeneous answers, disregarding differences in the input image.

(2) The SOTA closed-source LMM, GPT-4V, excels on task \ptloa, which focuses on querying the high-level semantics of the primitive concept (``what's in the image''). 
However, its performance significantly decreases on tasks requiring more precise low-level perception, \eg, \ptaoo and \ptlc.

(3) Fine-tuning the LLM backbone, Vicuna, with SVG inputs consistently outperforms fine-tuning the entire Llava model with image inputs. This highlights the potential of using SVG as an alternative representation in \imgtype.

(4) We note that SVG may inherently be inefficient in representing rendered 3D scenes and realistic images due to factors like camera perspectives, lighting, and shadows. While our focus in this work is on \imgtype, we leave the extension to other domains for future exploration.

\definecolor{bgblue}{HTML}{Cee7ee}
\definecolor{bgred}{HTML}{Eecece}

\begin{table*}[t]
\centering
\resizebox{\textwidth}{!}
{
\centering

\begin{tabular}{@{}llc|cccc}
    \toprule 
    & & \textbf{\begin{tabular}[c]{@{}c@{}} Input Type \end{tabular}} &
        \textbf{\begin{tabular}[c]{@{}c@{}} \ptloa \end{tabular}} &
        \textbf{\begin{tabular}[c]{@{}c@{}} \ptaoo \end{tabular}} &
        \textbf{\begin{tabular}[c]{@{}c@{}} \ptlc \end{tabular}} &
        \textbf{\begin{tabular}[c]{@{}c@{}} \ptcq \end{tabular}}\\
    \midrule
    \multirow{2}{*}{\textbf{Zero-Shot}}  & GPT-4V    & Image & \colorbox{bgred}{1.00} & 0.58 & 0.64 & \colorbox{bgred}{0.57} \\
                                & GPT-4     & SVG & 0.45 & 0.47 & 0.60 & 0.36 \\
    \midrule
    \multirow{3}{*}{\textbf{Finetuned}} & Llava-v1.5-7b  & Image & 0.50 & 0.50 & 0.50 & 0.45\\
                                & Vicuna     & SVG & \colorbox{bgblue}{0.93} & \colorbox{bgblue}{0.70} & \colorbox{bgblue}{0.99} & \colorbox{bgblue}{0.54} \\
    \bottomrule
\end{tabular}
}

\vspace{1mm}
\resizebox{\textwidth}{!}{
\begin{tabular}{@{}llc|ccccc}
    \toprule 
    & &  \multirow{2}{*}{\textbf{\begin{tabular}[c]{@{}c@{}} Input Type \end{tabular}}} &
            \multicolumn{2}{c}{\textbf{\begin{tabular}[c]{@{}c@{}} \ptsw \end{tabular}}} &
            \multicolumn{3}{c}{\textbf{\begin{tabular}[c]{@{}c@{}} \ptmaze \end{tabular}}} \\
    & & &
            \text{\begin{tabular}[c]{@{}c@{}} shape (acc$\boldsymbol{\uparrow}$) \end{tabular}} &
            \text{\begin{tabular}[c]{@{}c@{}} position (l2$\boldsymbol{\downarrow}$) \end{tabular}} &
            \text{\begin{tabular}[c]{@{}c@{}} connectivity (acc$\boldsymbol{\uparrow}$) \end{tabular}} &
            \text{\begin{tabular}[c]{@{}c@{}} start-pos (acc$\boldsymbol{\uparrow}$) \end{tabular}} &
            \text{\begin{tabular}[c]{@{}c@{}} end-pos (acc$\boldsymbol{\uparrow}$) \end{tabular}} \\
    \midrule
    \textbf{Zero-Shot}  & GPT-4V    & Image & \colorbox{bgred}{0.33} & 0.27 & 0.27 & \colorbox{bgred}{0.21} & \colorbox{bgred}{0.22} \\
    \midrule
    \multirow{2}{*}{\textbf{Finetuned}} & Llava-v1.5-7b   & Image & 0.04 & 0.67 & 0.26 & 0.03 & 0.03\\
                                & Vicuna    & SVG & \colorbox{bgblue}{0.15} & \colorbox{bgblue}{0.07} & \colorbox{bgblue}{0.54} & \colorbox{bgblue}{0.08} & \colorbox{bgblue}{0.09}\\
    \bottomrule
\end{tabular}
}
\vspace{-5pt}
\caption{Probing task results. We report the accuracy for the four question-answering tasks at the top. At the bottom, we use different metrics for different fields in the predicted scene description JSON. ``acc'' refers to accuracy (larger is better) while ``l2'' refers to the Euclidean distance between the predicted and ground truth [x, y] coordinates (lower is better).
Scores with a \colorbox{bgblue}{blue} background denote the better fine-tuned method compared to the SVG and Image representation. Scores with a \colorbox{bgred}{red} background denote tasks where fine-tuned methods cannot outperform zero-shot GPT-4V. Detailed analysis can be found in \S~\ref{subsec_app:svg_probing_exp}.}
\label{tab:probing_results}
\end{table*}

\begin{figure*}[t]
  \centering
  \includegraphics[width=0.6\textwidth]{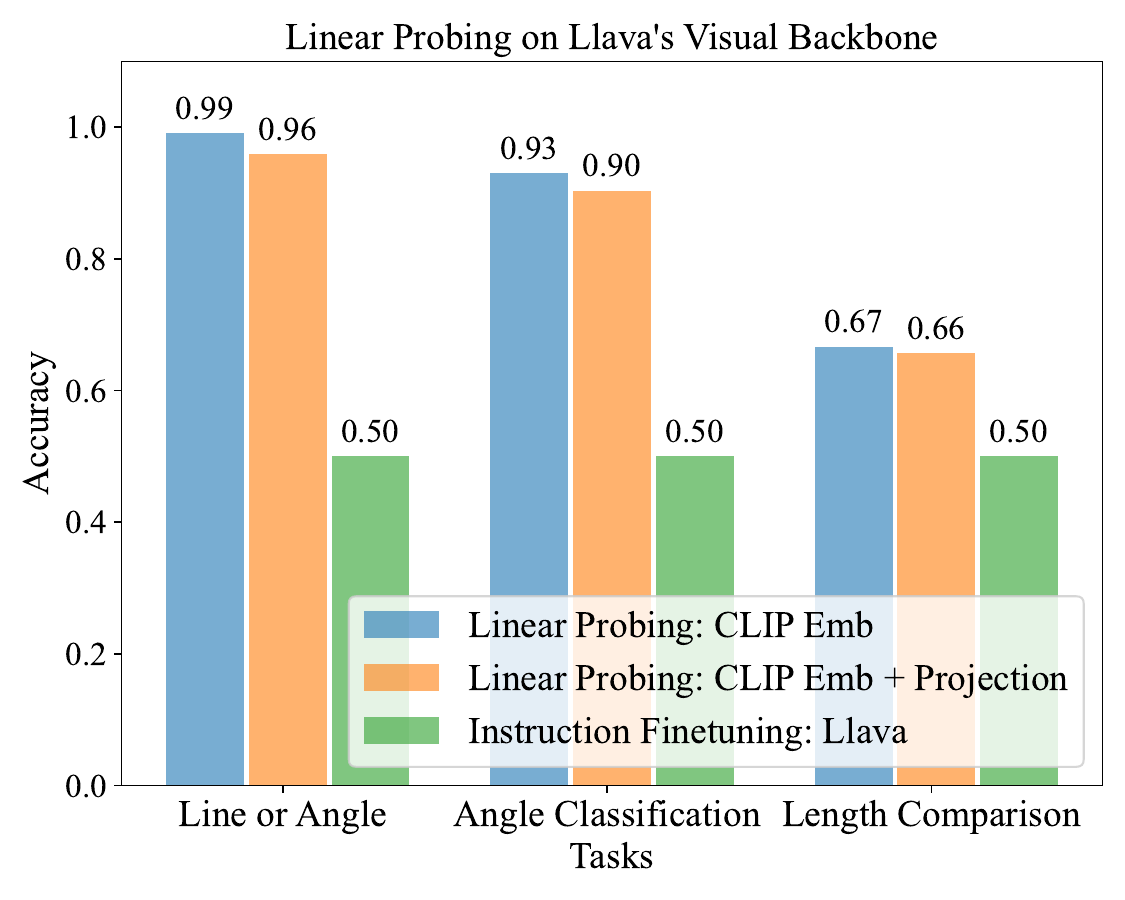}
  \caption{The average accuracy of linear probing, computed across ten epochs. Detailed training and testing scores for each epoch can be found in Figure~\ref{fig:linear_probing_training_details}.
  The results demonstrate that (1) CLIP embeddings are less effective for tasks requiring precise perception, such as \ptlc, in comparison to tasks that emphasize on higher-level semantics, such as \ptloa; (2) connecting to an LLM through the widely-used Llava-style architecture results in further diminished performance on tasks involving low-level visual details.}
  \label{fig:linear_probing_llava}
\end{figure*}
\definecolor{tabblue}{HTML}{1f77b4}
\definecolor{taborange}{HTML}{ff7f0e}
\begin{figure*}[t]
  \centering
  \includegraphics[width=\textwidth]{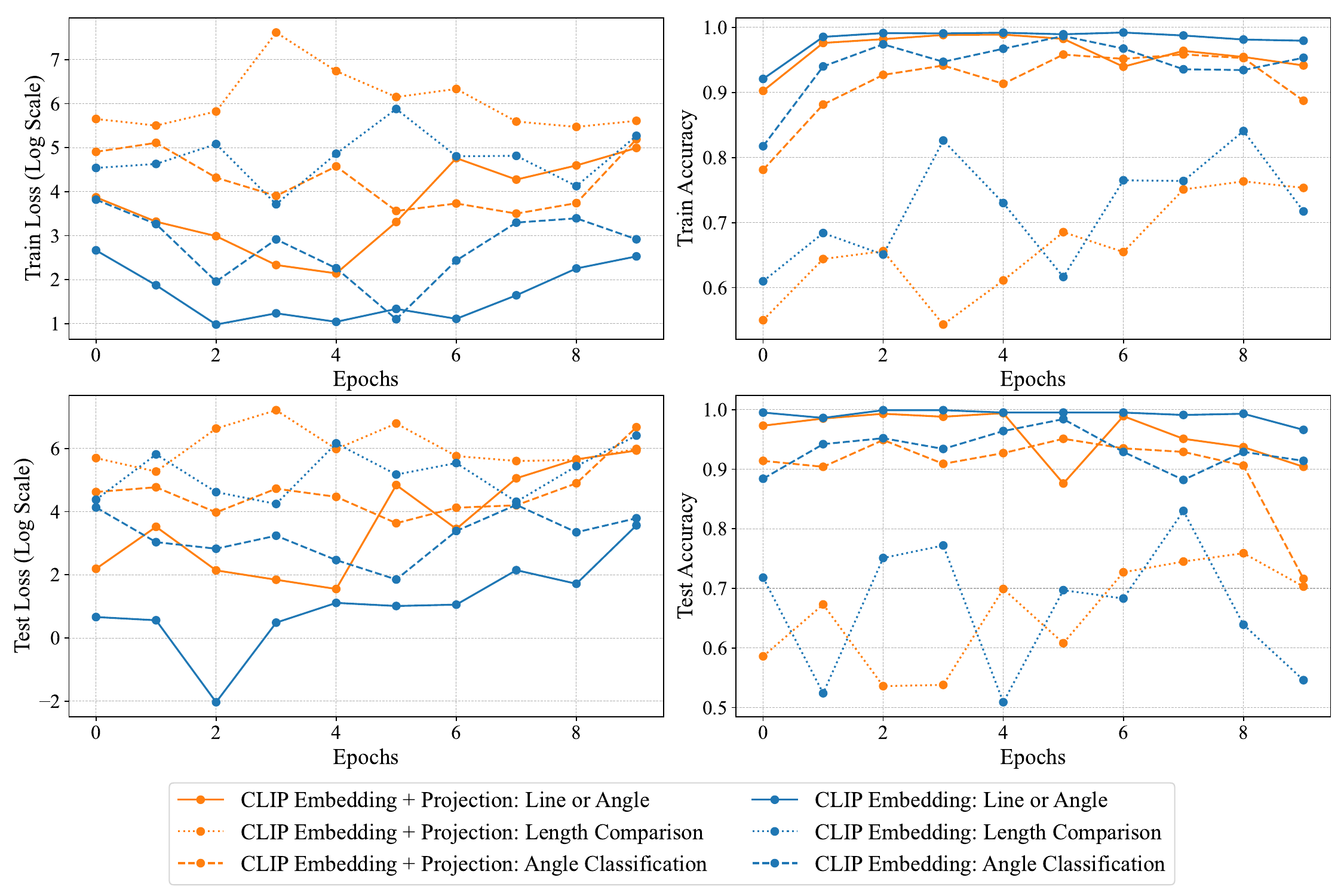}
  \caption{
Linear probing training details: Different line styles represent different tasks, while different colors refer to different visual embeddings used for training the linear classifier. The training loss (top-left) shows that the projected embedding (\textcolor{taborange}{orange lines}) learns at a slower pace compared to the original CLIP embedding (\textcolor{tabblue}{blue lines}). The training accuracy (top-right) reveals that for certain tasks, such as \ptlc, the model continues to struggle with overfitting the training set even after 10 epochs.}

  \label{fig:linear_probing_training_details}
\end{figure*}

\subsection{Llava's Failure Mode in Visual Reasoning with Vector Graphics}
\label{subsec_app:linear_probing}
We further investigate whether the difficulty in understanding low-level visual features of Llava models stems from (1) the visual backbone itself, i.e., CLIP, or (2) the bridge between the visual backbone and the LLM backbone. We include a set of \textbf{Linear Probing} experiments on three binary classification probing tasks, where we train a simple linear classifier based on the visual backbone features (before and after projection) of the Llava model. As shown in Figure~\ref{fig:linear_probing_llava}:

(1) In tasks requiring more precise low-level perception, such as \ptaoo and \ptlc, CLIP embeddings are inherently less effective at capturing relevant features. Furthermore, as shown in Figure~\ref{fig:linear_probing_training_details}, in some tasks, e.g., \ptlc, linear regression even fails to achieve 90\%+ training accuracy after 10 epochs of training, struggling to converge.

(2) When connected to an LLM using the projection layer, the visual features in Llava become less effective for low-level visual reasoning. Additionally, there is a significant gap between linear probing and instruction fine-tuning performance. These results suggest that even if the backbone does preserve useful features, the LLM cannot effectively leverage those visual tokens after projection.

We hypothesize that the failure mode likely stems from the multimodal pretraining and instruction-tuning paradigm, where the tasks are biased towards high-level semantics, such as image captioning~\citep{mscoco, sidorov2020textcaps} and natural-image-based VQA~\citep{vqav2, visual_genome, okvqa, aokvqa}.
The training mixtures~\citep{llava, llava15, dai2023instructblip, chen2023sharegpt4v} for current LMMs predominantly focus on high-level features of images, providing little incentive for models to retain low-level visual details. For example, the caption of an image containing a 2D maze, such as the one shown in Figure~\ref{fig:overview}, is likely to be ``A 2$\times$2 maze with black lines, a red circle and a star.'' and may not include detailed configurations of the mazes, such as the precise locations of the walls, the red circle, and the red star.

\subsection{Remaining Challenges of Using SVG Representations}
\label{subsec_app:remaining_challenges_of_raw_svg_repre}

Although we have demonstrated that SVG can serve as a promising alternative representation for reasoning about \imgtype, we identify several remaining challenges:

(1) Pretrained LLMs, including the most capable ones such as GPT-4~\citep{openai2023gpt4}, possess limited out-of-the-box understanding of SVG code. This limitation is evidenced by the low zero-shot performance of GPT-4 with SVG input (see row 2 in Table~\ref{tab:probing_results}).

(2) Even after finetuning, the SVG-based LLM may still underperform zero-shot GPT-4V on certain tasks, particularly those involving complex scenes, such as \ptsw and \ptmaze. In these instances, the SVG code becomes excessively verbose. These findings suggest that learning a model to directly comprehend the raw SVG code of an entire image poses significant challenges. 

(3) A fundamental challenge, irrespective of the chosen representation for visual input, is the lack of generalization capability to unseen tasks and various \imgtype image domains. If we rely on existing LMM training mixtures, even any image can be converted into SVG code, the tasks remain biased towards high-level semantics. In addition, it is infeasible to directly manually construct and annotate $\langle$SVG, question, answer$\rangle$ pairs covering diverse tasks with \imgtype.

These challenges motivated us to propose another layer of abstraction, the \bl{}, aimed at bridging the gap between low-level perception and high-level language reasoning on downstream tasks.

\section{Error Analysis Details}
\label{sec_app:error_analysis_details}
\begin{figure*}[t]
  \centering
  \begin{subfigure}[c]{0.72\textwidth}
    \includegraphics[width=\textwidth]{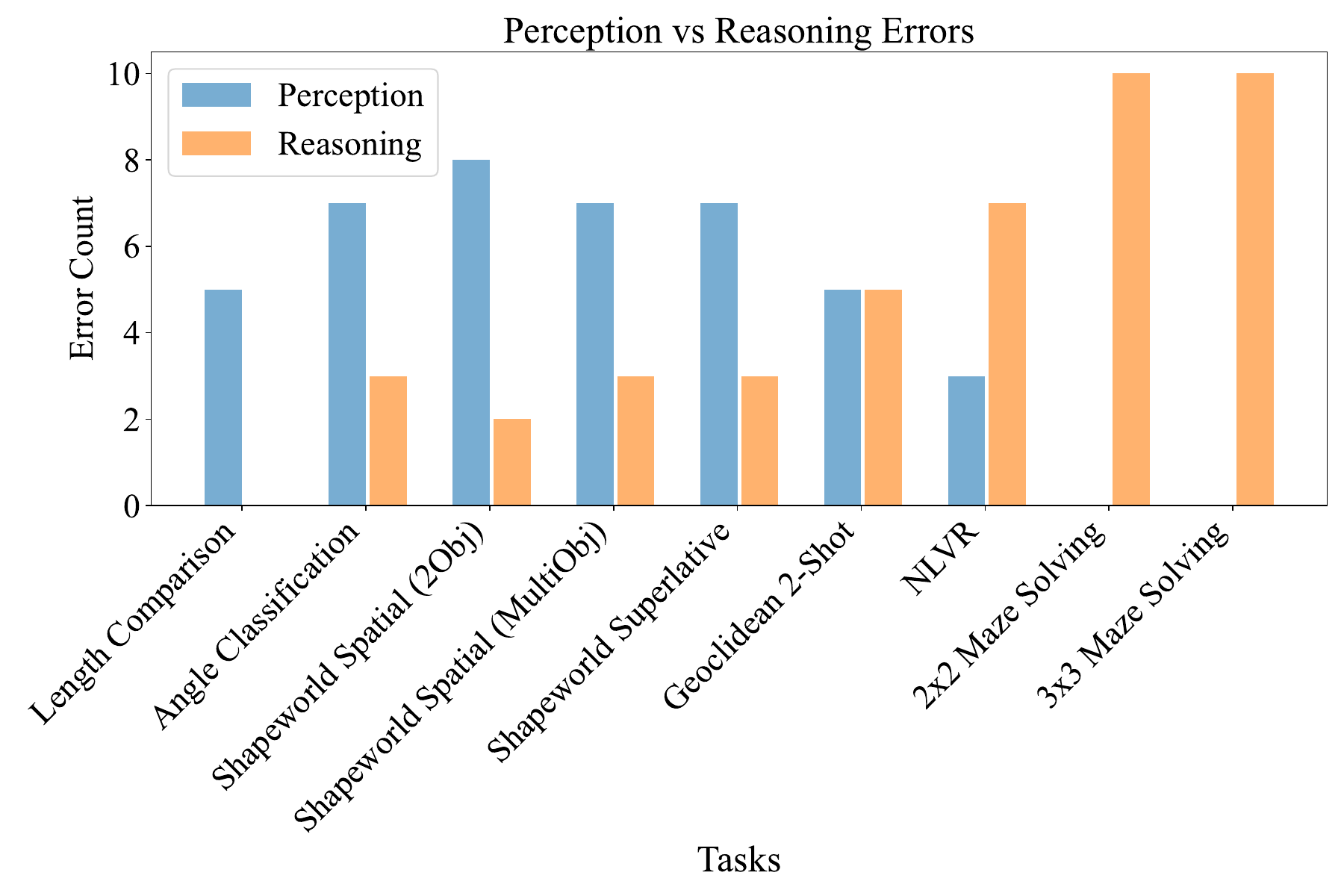}
    \label{fig:error_single_task}
  \end{subfigure}
  \hspace{-5pt}
  \hfill
  \begin{subfigure}[c]{0.24\textwidth}
    \includegraphics[width=\textwidth]{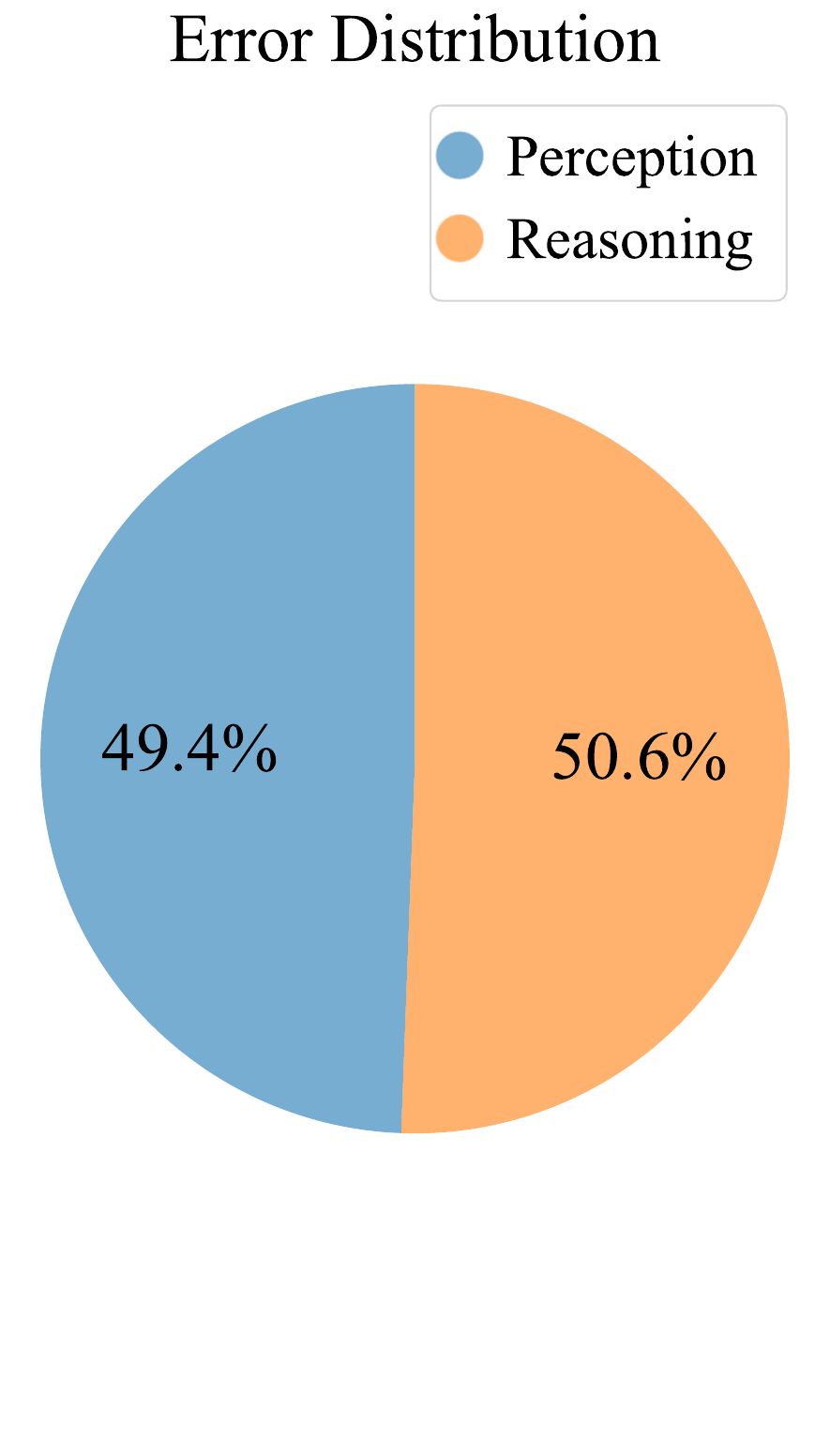}
    \label{fig:error_all_task}
  \end{subfigure}
  \vspace{-10pt}
  \caption{Error distribution by \ourstxt between perception and reasoning on low-level \imgtype reasoning.}
  \label{fig:error_categorization}
\end{figure*}
As introduced in \S~\ref{sec:method}, the proposed \ours{} consists of two stages focused on perception—namely, Image-to-SVG and SVG-to-\blabbr{}, and one stage focused on reasoning, i.e., \blabbr{}-to-final answer. 
We aim to investigate the errors in both the perception and reasoning modules.

For each task, we manually examine 10 error cases and determine whether the error primarily stems from the perception stage or the reasoning stage. We task a human with reviewing the reconstructed image from the \blabbr{} representation and assessing the question of the task instance. 
If, for a human, the reconstructed image is still insufficient for solving the task, we classify this error as a perception error. Otherwise, it is categorized as a reasoning error. Figure~\ref{fig:error_categorization} illustrates the distribution of errors between perception and reasoning stages. 
We further identify some typical categories of perception and reasoning errors as follows:

\begin{figure*}[t]
  \centering
  \includegraphics[width=\textwidth]{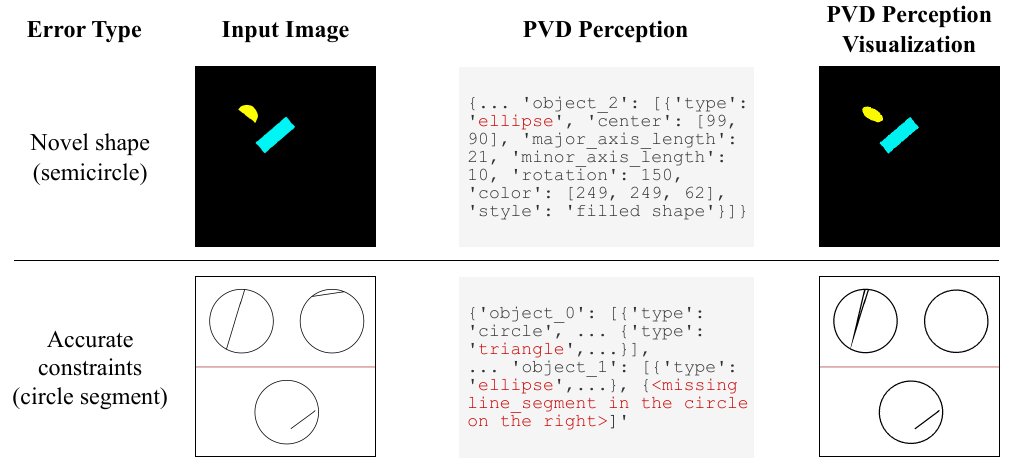}
  \caption{Perception error examples. The example at the top illustrates an error wherein the SVG-to-\blabbr model predicts a semicircle as an ellipse. The example at the bottom demonstrates that the SVG-to-\blabbr model struggles to decode overlapping primitives with accurate constraints, such as a segment of a circle.}
  \label{fig:perception_error_examples}
\end{figure*}
\paragraph{Common perception errors.}
(1) \textbf{Novel shapes not covered by the \bl{} (\blabbr{})}: For example, as visualized in Figure~\ref{fig:perception_error_examples}, the Shapeworld dataset includes a ``semicircle'' shape type which is not in the \blabbr{} ontology; we see that the learned SVG-to-\blabbr{} model tends to predict it as an ellipse. This perception error directly contributes to the inferior performance of \oursmm compared to GPT-4o on the Shapeworld tasks, as shown in Table~\ref{tab:main_result}.

(2) \textbf{Accurate constraints between primitives}: Although the \blabbr{} accommodates scenarios where multiple objects of the same color overlap, the attributes, e.g., position, of each object are decided independently and randomly. Thus, the SVG-to-\blabbr{} model often fails to capture intentional constraints between objects; for example, a line that perfectly segments a circle. These constraints are particularly emphasized in the \dtgeo task (Figure~\ref{fig:perception_error_examples}), where \oursfull{} struggles to outperform GPT-4V and GPT-4o. 

(3) \textbf{Very small objects}: During inference, the iterative decomposition process heuristically ignores SVG paths that only contribute only minor differences to the reconstructed image. This method effectively reduces noise from the rule-based image-to-SVG converter but may omit very small objects in some cases. Adjusting this threshold is necessary for specific scenarios.

\begin{figure*}[t]
  \centering
  \includegraphics[width=\textwidth]{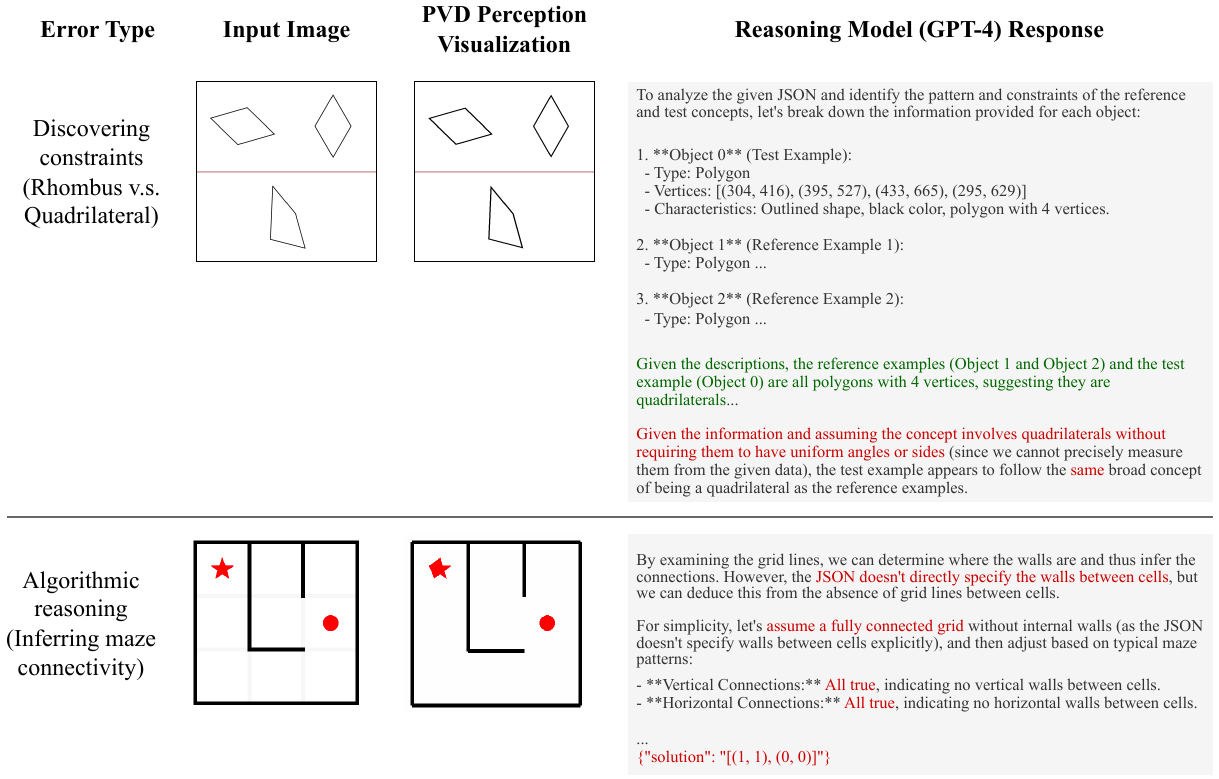}
  \caption{Reasoning error examples: The example at the top demonstrates that the reasoning model fails to uncover the deeper constraints within the perceived polygons. It is able to recognize that they are all quadrilaterals but unable to further discern that the reference concepts are rhombuses with four equal sides, while the test concept is not. The example at the bottom illustrates that the reasoning model struggles to infer connectivity based on the perceived grid, thus failing to provide the correct solution.}
  \label{fig:reasoning_error_examples}
\end{figure*}
\paragraph{Common reasoning errors.}

(1) \textbf{Discovering intentional constraints}: Without specific queries, the reasoning model can fail to identify intentional constraints. For example, differentiating a rhombus from a general quadrilateral, as shown in Figure~\ref{fig:reasoning_error_examples}.

(2) \textbf{Handling ambiguity}: Visual inputs sometimes provide useful inductive biases that can help the model better understand the task or make reasonable assumptions when the instructions are ambiguous. For instance, when presenting an angle in an image and asking whether it is an acute or obtuse angle, as in Figure~\ref{fig:downstream_tasks}, it is visually straightforward to assume that the angle is defined by the middle point as the vertex with rays extending outwards. However, without such visual cues, reasoning over pure symbolic representations makes it challenging to infer which angle the question refers to among the detected undirected edges. To mitigate ambiguity, adding more precise instructions for \ourstxt{} is necessary in some tasks.

(3) \textbf{Algorithmic reasoning}: Language-based reasoners can struggle with complex multi-step reasoning tasks, such as inferring the connectivity (Figure~\ref{fig:reasoning_error_examples}) of a maze using the vertices and edges of the grid in pixel coordinates, or counting the number of objects located within a certain box.

\section{\bl{} (\blabbr{}) Details}
\label{sec_app:base_layer_details}

\begin{figure*}[t]
  \centering
  \includegraphics[width=\textwidth]{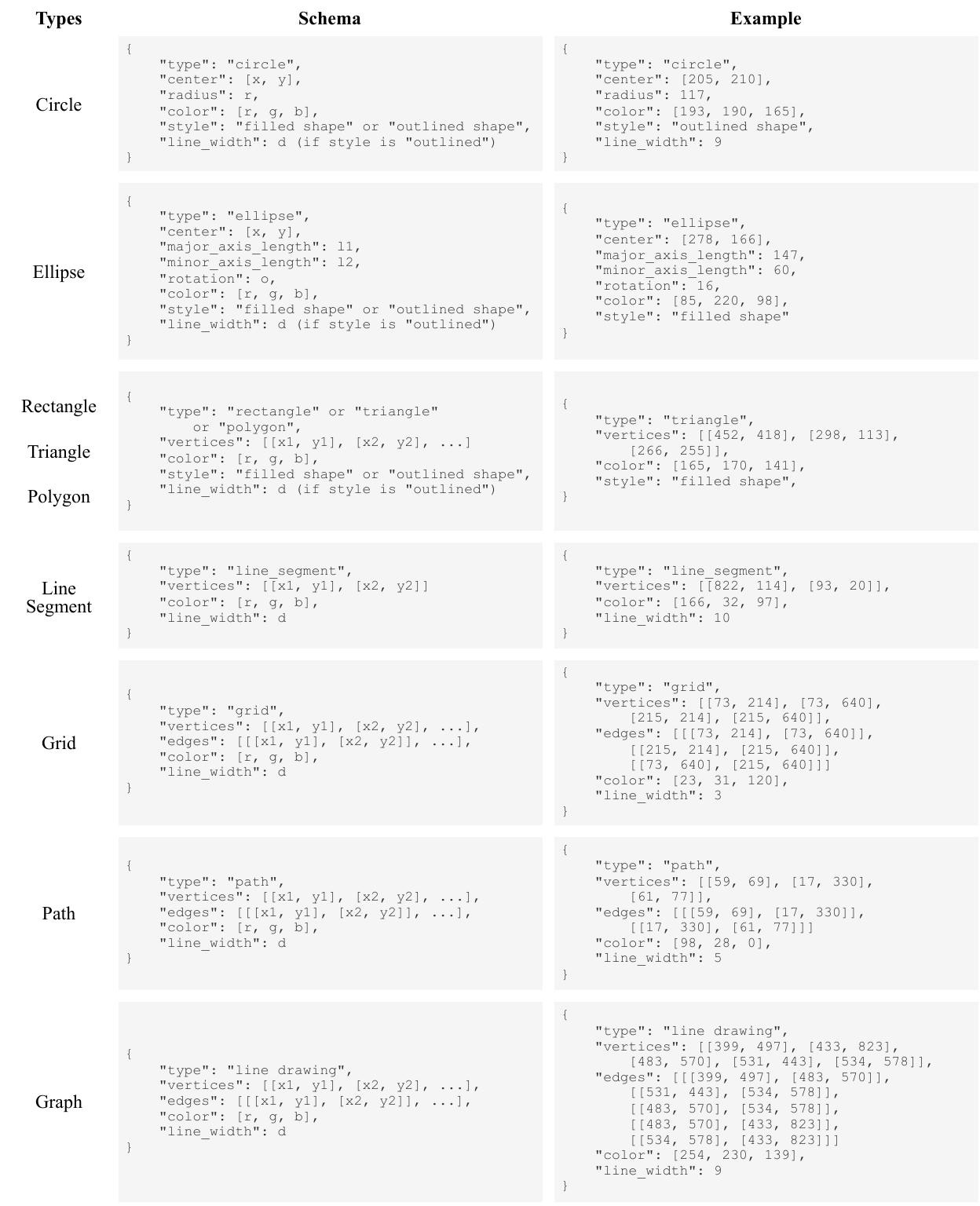}
  \caption{\blabbr{} JSON schema definition.}
  \vspace{-10pt}
  \label{fig:pvd_json_schema}
\end{figure*}
\paragraph{\blabbr{} JSON schema definition:} See Figure~\ref{fig:pvd_json_schema}.

\begin{table}[t]
\centering
\resizebox{0.48\textwidth}{!}
{
\centering
\begin{tabular}{@{}lllc}
    \toprule 
    & \textbf{Style} & \textbf{Concept} & \textbf{\begin{tabular}[c]{@{}c@{}} \# Instances \end{tabular}} \\
    \midrule
    \multirow{9}{*}{\textbf{\begin{tabular}[c]{@{}l@{}} Single Object \end{tabular}}} &  
    \multirow{9}{*}{\begin{tabular}[c]{@{}l@{}} Filled\\or\\Outlined\end{tabular}} & Circle & 10K \\
    & & Ellipse & 10K \\ 
    & & Rectangle & 10K \\ 
    & & Triangle & 10K \\ 
    & & Polygon & 20K \\ 
    & & Line Segment & 10K \\ 
    & & Grid & 10K \\ 
    & & Path & 10K \\ 
    & & Graph & 10K \\ 
    \midrule
    \multirow{8}{*}{\textbf{\begin{tabular}[c]{@{}l@{}} Composition \end{tabular}}} & 
    \multirow{4}{*}{\begin{tabular}[c]{@{}l@{}} Filled \end{tabular}} & Circle & 5K \\
    & & Rectangle & 5K \\
    & & Triangle & 5K \\
    & & Line Segment & 5K \\
    \cmidrule(lr){2-4}
    & \multirow{4}{*}{\begin{tabular}[c]{@{}l@{}} Outlined \end{tabular}} & Circle & 10K\\
    & & Rectangle & 10K \\
    & & Triangle & 10K \\
    & & Line Segment & 10K \\
    \midrule
    & & \textbf{Total} & 160K \\
    \bottomrule
\end{tabular}
}
\caption{\blabbr{} 160K dataset statistics.}

\label{tab:base_layer_detail}
\end{table}
\paragraph{Generation procedures (Single Object):}

\begin{itemize}
\item \textbf{Circle:} Randomly sample a center and a radius to draw a circle within the canvas.
\item \textbf{Ellipse:} Randomly sample a center, a major axis, and a minor axis, then randomly rotate by an angle. Verify if the ellipse is largely within the canvas; if not, try again.
\item \textbf{Rectangle:} Randomly sample a top-left corner, a width, and a height, then randomly rotate by an angle. Verify if the rectangle is largely within the canvas; if not, try again.
\item \textbf{Triangle:} Randomly sample three points as vertices to draw a triangle. Check if the area is larger than a threshold; if not, try again.
\item \textbf{Polygon:} Randomly sample $N \in [5, 20]$ points. Order the points with respect to the centroid so that no intersections will happen when connected with a polyline. Draw a polygon with the sampled points. Check if the polygon has an area larger than a threshold; if not, try again.
\item \textbf{Path:} Randomly and iteratively sample $N \in [3, 16]$ points, connect the newly sampled point with the previous point to form a line segment. Verify if the newly added line segment does not intersect with any of the previous line segments; if yes, resample the point.
\item \textbf{Grid:} Sample a grid of points with a size $M \times N$ where $M, N \in [2, 6]$. First, use Depth First Search (DFS) algorithm to connect all grid vertices into a connected graph. Then randomly add additional edges between adjacent vertices.
\item \textbf{Graph:} Randomly sample $N \in [4, 16]$ points. First, use Kruskal's algorithm~\citep{kruskal1956shortest} to find a Minimum Spanning Tree that connects all the points. Then randomly add additional edges to the graph.
\end{itemize}

\paragraph{Generation procedures (Composition):}
Iteratively draw shapes on the canvas chosen from the following set of object types: [``circle'', ``rectangle'', ``triangle'', ``line segment'']. After the first shape is drawn, at each iteration, the later shapes are constrained to have the same color as the previous shapes. We ensure overlap between the newly added shape and the previous shapes, while making sure that the intersection ratio does not exceed a predefined threshold. This prevents cases where one shape entirely contains another, making it impossible to decode into individual \bl{} elements.

\paragraph{\blabbr{} 160K dataset:}
Using the aforementioned generation procedure, we generate a large-scale dataset containing 160K $\langle$SVG, \blabbr{}$\rangle$ pairs for training the LLM-based SVG-to-\blabbr{} model. The detailed configuration can be found in Table~\ref{tab:base_layer_detail}.

\begin{table}[t]
\centering
\resizebox{0.5\textwidth}{!}
{
\centering
\begin{tabular}{@{}lccc}
    \toprule 
    & \textbf{\begin{tabular}[c]{@{}c@{}} SSIM \end{tabular}} 
    & \textbf{\begin{tabular}[c]{@{}c@{}} DINOv2 Score\end{tabular}} 
    & \textbf{\begin{tabular}[c]{@{}c@{}} CLIP Score\end{tabular}}
    \\
    \midrule
    w/o aug & 0.892& 0.874& 0.886 \\
    w/ aug & \textbf{0.895}& \textbf{0.893}& \textbf{0.893} \\
    \bottomrule
\end{tabular}
}
\caption{Impact of the data augmentation (Gaussian Blur and Pixel Noise detailed in \S\ref{subsec:svg-to-baselayer}) on SVG-to-\blabbr{} model perception performance.}
\vspace{-10pt}
\label{tab:impact_aug}
\end{table}

\paragraph{Data augmentation details:}
To enhance the robustness of the SVG-to-\blabbr{} model to images with various sizes and quality, we introduce the following randomized data augmentation during data generation.
\begin{itemize}
\item \textbf{Random pixel noise}: Probability (how often to apply the augmentation): 0.1; Ratio range (what portion of the selected area will be filled with noise pixels): (0.01, 0.05); Intensity range (the intensity of the noise pixels): (0.1, 1.0); Dilate range (how many pixels will the selection area be extended from the boundary): (1, 3) in pixels; Noise size: (1, 3) in pixels.
\item \textbf{Gaussian blur}: Probability (how often to apply the augmentation): 0.1; Radius: (0.1, 0.5).
\end{itemize}

Table~\ref{tab:impact_aug} shows the ablation study with and without the data augmentations.

\section{Image-to-SVG Visual Encoding Quality}
\label{app:tmlr_added_vtracer_encoding_quality}

\paragraph{VTracer Encoding Quality.}
As introduced in Section~\ref{subsec:image-to-svg}, the first step of our proposed perception module is to encode a raster image into an SVG representation using VTracer, a rule-based converter that we found yields near-perfect reconstructions for vector-graphic–style inputs. To verify this, we convert the VTracer-encoded SVGs back into PNG images and compare them with the originals. The results are presented in Table~\ref{tab:vtracer_reconstruction}.

\begin{table}[ht]
  \centering
  \resizebox{0.56\textwidth}{!}
{
  \begin{tabular}{@{}lccc@{}}
    \toprule
    Task                          & SSIM  & DINOv2 Score & CLIP Score \\
    \midrule
    Angle Classification          & 0.990 & 0.991        & 0.997      \\
    Length Comparison             & 0.994 & 0.970        & 0.986      \\
    Shapeworld Spatial 2Obj       & 0.997 & 0.984        & 0.995      \\
    Shapeworld Spatial mObj       & 0.991 & 0.984        & 0.988      \\
    Shapeworld Superlative        & 0.991 & 0.982        & 0.987      \\
    NLVR                          & 0.994 & 0.989        & 0.993      \\
    Geoclidean                    & 0.648 & 0.994        & 0.995      \\
    Maze 2×2                      & 0.998 & 0.906        & 0.775      \\
    Maze 3×3                      & 0.998 & 0.998        & 0.989      \\
    \midrule
    \textbf{ALL}                  & 0.956 & 0.978        & 0.967      \\
    \bottomrule
  \end{tabular}
}
\vspace{-5pt}
  \caption{VTracer SVG encoding reconstruction metrics. The exceptionally low SSIM score on Geoclidean is due to the fact that most pixels are transparent. Since we compute SSIM only on non-transparent regions, the score becomes more sensitive to slight pixel differences. However, the reconstruction appears visually near-perfect across all tasks, as reflected by the high DINOv2 score.
}
\vspace{-5pt}
  \label{tab:vtracer_reconstruction}
\end{table}

We observe near-perfect reconstruction quality from VTracer.  For reference, the average reconstruction quality from the PVD representation (after SVG-to-PVD conversion) is SSIM = 0.895, DINOv2 Score = 0.880, and CLIP Score = 0.893, indicating that most perception errors arise during the SVG-to-PVD step. However, as discussed in Section~\ref{subsec:svg-to-baselayer}, although SVG preserves all low-level features, it is extremely verbose and noisy—making it unintelligible to LLMs and LMMs. This observation motivates the development of the intermediate PVD representation.

\begin{figure*}[t]
  \centering
  \includegraphics[width=0.95\textwidth]{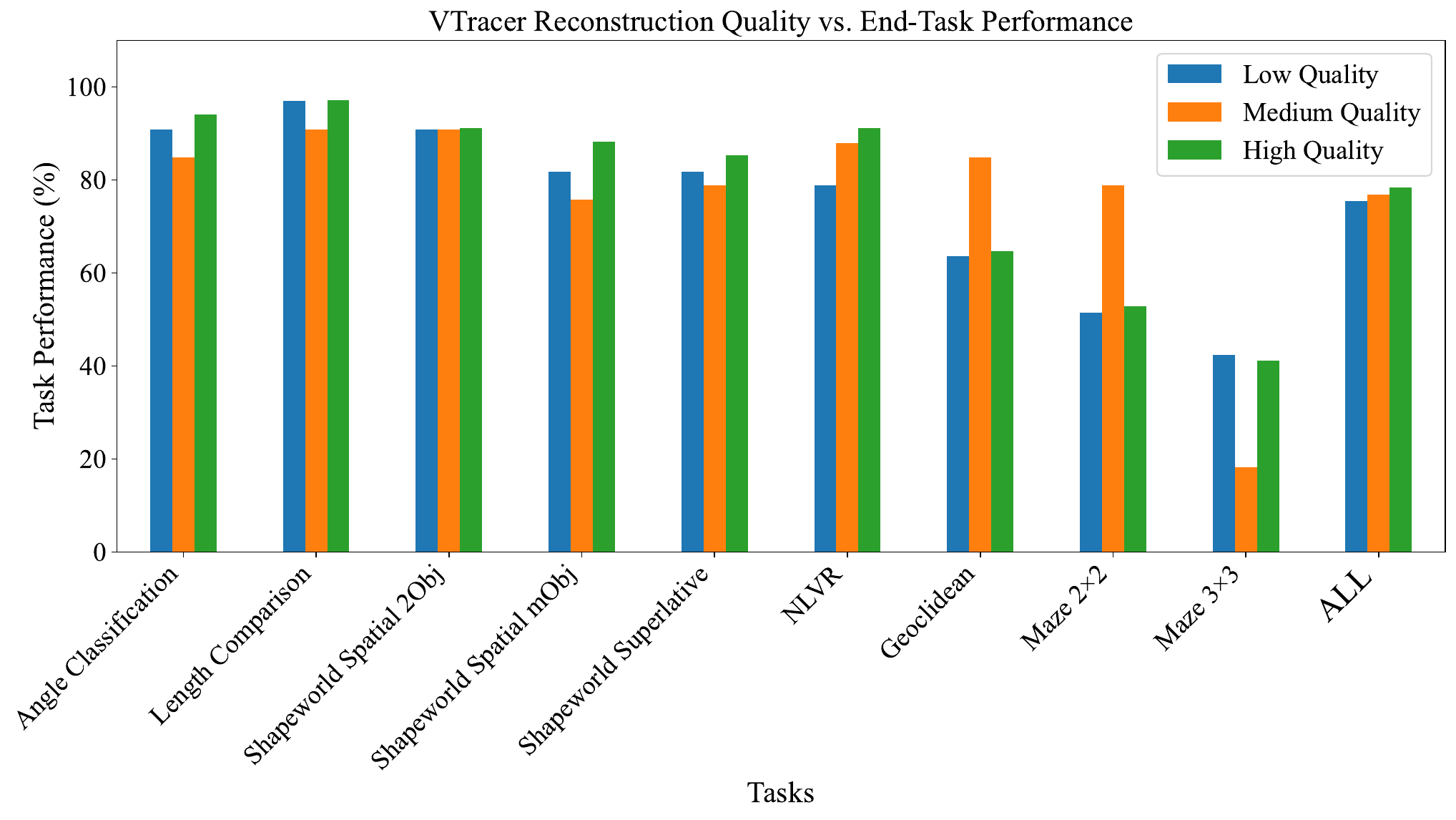}
  \vspace{-5pt}
  \caption{Impact of SVG encoding quality on end-task performance.}
  \vspace{-10pt}
  \label{fig:vtracer_rec_vs_end_task}
\end{figure*}

\paragraph{Impact of VTracer Error to the End-task Performance.} 
To further investigate whether imperfections in SVG encoding significantly affect end-task performance. We sort the instances for each task with respect to VTracer reconstruction quality, using the aggregated metric: SSIM + DINOv2 score + CLIP score. The instances are then grouped into three bins corresponding to low, medium, and high VTracer quality. We examine the end-task performance (VDLM-mm + GPT-4o) of these groups to investigate whether performance is sensitive to VTracer errors. In Figure~\ref{fig:vtracer_rec_vs_end_task}, we show the average accuracy on each task for each bin. Note that each bin has about 33 instances.

Overall, we observe that end-task performance is fairly robust to vectorization errors, while also exhibiting a trend similar to what we reported in Section~\ref{subsec:analysis_perception_impact}, i.e., a positive correlation between perception quality and end-task performance.

\section{Additional Ablations on PVD Perception}
\label{app:tmlr_added_llm_choice_and_png_to_pvd}

\begin{table}[ht]
\centering
\resizebox{0.68\textwidth}{!}
{
\begin{tabular}{lccc}
\toprule
\textbf{PVD Model }& \textbf{SSIM ↑} & \textbf{DINOv2 Score ↑} & \textbf{CLIP Score ↑} \\
\midrule
SVG‐to‐PVD (Mistral‐7B) & 0.895 & 0.880 & 0.893 \\
SVG‐to‐PVD (Qwen2.5‐7B) & 0.889 & 0.876 & 0.880 \\
\bottomrule
\end{tabular}
}
\vspace{-5pt}
\caption{Perception metrics of different SVG-to-PVD models.}
\label{tab:svg_to_pvd_perception}
\vspace{-5pt}
\end{table}
\vspace{-5pt}
\begin{table*}[ht]
\centering
\resizebox{0.90\textwidth}{!}
{
\centering

\begin{tabular}{@{}ll|ccccccccc|c}
    \toprule
    \textbf{\begin{tabular}[c]{@{}c@{}} Reasoner \end{tabular}} & 
    \textbf{\begin{tabular}[c]{@{}c@{}} PVD\\Model \end{tabular}} &
    \textbf{\begin{tabular}[c]{@{}c@{}} AC \end{tabular}} &
    \textbf{\begin{tabular}[c]{@{}c@{}} LC \end{tabular}} &
    \textbf{\begin{tabular}[c]{@{}c@{}} SW-S\\2Obj \end{tabular}} &
    \textbf{\begin{tabular}[c]{@{}c@{}} SW-S\\mObj \end{tabular}}&
    \textbf{\begin{tabular}[c]{@{}c@{}} SW\\Sup \end{tabular}} &
    \textbf{\begin{tabular}[c]{@{}c@{}} NLVR \end{tabular}} &
    \textbf{\begin{tabular}[c]{@{}c@{}} Geo \end{tabular}} &
    \textbf{\begin{tabular}[c]{@{}c@{}} Maze\\2$\times$2 \end{tabular}} & 
    \textbf{\begin{tabular}[c]{@{}c@{}} Maze\\3$\times$3 \end{tabular}} & 
    \textbf{\begin{tabular}[c]{@{}c@{}} All \end{tabular}} \\
    \midrule
    GPT‐4o   & –                 & 0.63 & 0.57 & 0.97      & 0.82      & 0.92   & 0.81 & 0.71 & 0.46      & 0.08      & 0.663 \\
GPT‐4o   & Mistral‐7B        & 0.90 & 0.95 & 0.91      & 0.82      & 0.82   & 0.86 & 0.71 & 0.61      & 0.34      & 0.769 \\
GPT‐4o   & Qwen2.5‐7B        & 0.82 & 0.98 & 0.99      & 0.79      & 0.89   & 0.83 & 0.68 & 0.70      & 0.29      & 0.774 \\
    \bottomrule
\end{tabular}
}
\vspace{-5pt}
\caption{End-task performance with different SVG-to-PVD models.}
\vspace{-10pt}
\label{tab:qwen_pvd_end_task}
\end{table*}

\paragraph{Different LLM choices for SVG-to-PVD model.}
We investigate the effect of using a different LLM for SVG-to-PVD translation. Specifically, we train a Qwen-2.5-7B\footnote{\url{https://huggingface.co/Qwen/Qwen2.5-7B}} base model on the PVD-160k dataset using the same number of epochs as our original model. We observe comparable perception and end-task performance to our original PVD model, which is based on Mistral-7B. Table~\ref{tab:svg_to_pvd_perception} and Table~\ref{tab:qwen_pvd_end_task} report the perception scores and end-task performance of SVG-to-PVD models trained with different LLM backbones. These results suggest that the SVG-to-PVD translation process is robust to the choice of LLM.

\begin{table}[ht]
\centering
\resizebox{0.59\textwidth}{!}
{
\begin{tabular}{lcc}
\toprule
\textbf{PVD Model} &\textbf{ Training Epochs }& \textbf{Final Loss ↓} \\
\midrule
SVG‐to‐PVD (Qwen2.5‐7B)       & 3 epochs & 0.052 \\
SVG‐to‐PVD (Mistral‐7B)      & 3 epochs & 0.051 \\
\midrule
PNG‐to‐PVD (Qwen2.5‐VL-7B)    & 3 epochs & 0.243 \\
\bottomrule
\end{tabular}
}
\vspace{-5pt}
\caption{Comparison of training loss between the PNG-to-PVD and SVG-to-PVD models.}
\label{tab:pvd_training_loss}
\vspace{-5pt}
\end{table}

\begin{table}[ht]
\centering
\resizebox{0.68\textwidth}{!}
{
\begin{tabular}{lccc}
\toprule
\textbf{PVD Model }& \textbf{SSIM ↑} & \textbf{DINOv2 Score ↑} & \textbf{CLIP Score ↑} \\
\midrule
SVG‐to‐PVD (Mistral‐7B) & 0.895 & 0.880 & 0.893 \\
SVG‐to‐PVD (Qwen2.5‐7B) & 0.889 & 0.876 & 0.880 \\
\midrule
PNG‐to‐PVD (Qwen2.5‐VL-7B) & 0.262 & 0.320 & 0.385 \\
\bottomrule
\end{tabular}
}
\vspace{-5pt}
\caption{Comparison of perception metrics between the PNG-to-PVD and SVG-to-PVD models.}
\label{tab:png_to_pvd_perception}
\vspace{-5pt}
\end{table}
\paragraph{Ablation with PNG-to-PVD model.}
To further validate the design choice of using SVG for initial visual encoding, we compare it with directly training a PNG-to-PVD model using a large multimodal model. Specifically, we train Qwen2.5-VL-7B~\footnote{\url{https://huggingface.co/Qwen/Qwen2.5-VL-7B-Instruct}} on the same PVD-160k dataset, where the inputs are png images and the target outputs are PVD strings. 
Table~\ref{tab:pvd_training_loss} and Table~\ref{tab:png_to_pvd_perception} show the comparison between the PNG-to-PVD and SVG-to-PVD models in terms of final training loss and perception metrics.
Using identical hyperparameters, we find that directly translating PNG to PVD is significantly less effective, as evidenced by a much higher loss and substantially lower perception performance. We observe that the PNG-to-PVD model rarely produces valid PVD JSON outputs and yields significantly worse reconstruction performance (entirely not usable for downstream reasoning).

\section{Additional Experiments Using Open-Source LMMs as Reasoners}
\label{app:tmlr_added_open_source_lmm_reasoner}
\vspace{-5pt}
\begin{table*}[ht]
\centering
\resizebox{0.95\textwidth}{!}
{
\centering

\begin{tabular}{@{}ll|ccccccccc|c}
    \toprule
    \textbf{\begin{tabular}[c]{@{}c@{}} Reasoner \end{tabular}} & 
    \textbf{\begin{tabular}[c]{@{}c@{}} PVD\\Model \end{tabular}} &
    \textbf{\begin{tabular}[c]{@{}c@{}} AC \end{tabular}} &
    \textbf{\begin{tabular}[c]{@{}c@{}} LC \end{tabular}} &
    \textbf{\begin{tabular}[c]{@{}c@{}} SW-S\\2Obj \end{tabular}} &
    \textbf{\begin{tabular}[c]{@{}c@{}} SW-S\\mObj \end{tabular}}&
    \textbf{\begin{tabular}[c]{@{}c@{}} SW\\Sup \end{tabular}} &
    \textbf{\begin{tabular}[c]{@{}c@{}} NLVR \end{tabular}} &
    \textbf{\begin{tabular}[c]{@{}c@{}} Geo \end{tabular}} &
    \textbf{\begin{tabular}[c]{@{}c@{}} Maze\\2$\times$2 \end{tabular}} & 
    \textbf{\begin{tabular}[c]{@{}c@{}} Maze\\3$\times$3 \end{tabular}} & 
    \textbf{\begin{tabular}[c]{@{}c@{}} All \end{tabular}} \\
    \midrule
    Qwen2.5-VL-72B  & –                & 0.78 & 0.75 & 0.98 & 0.75 & 0.92 & 0.72 & 0.67 & 0.64 & 0.15 & 0.707 \\
    Qwen2.5-VL-72B  & Mistral‐7B       & 0.59 & 0.96 & 0.96 & 0.78 & 0.89 & 0.69 & 0.73 & 0.66 & 0.28 & 0.727 \\
    \bottomrule
\end{tabular}
}
\vspace{-5pt}
\caption{End-task performance with open-source LMM reasoners.}
\vspace{-10pt}
\label{tab:open_source_lmm}
\end{table*}
We added new experimental results demonstrating that VDLM can also work effectively with recent open-source LMMs. Specifically, we use Qwen-2.5-VL-72B\footnote{\url{https://huggingface.co/Qwen/Qwen2.5-VL-72B-Instruct}} as the reasoner to compare performance with and without the inclusion of PVD representations. Across 9 tasks, we observe a 2\% overall improvement.

\begin{figure*}[t]
  \centering
  \includegraphics[width=0.8\textwidth]{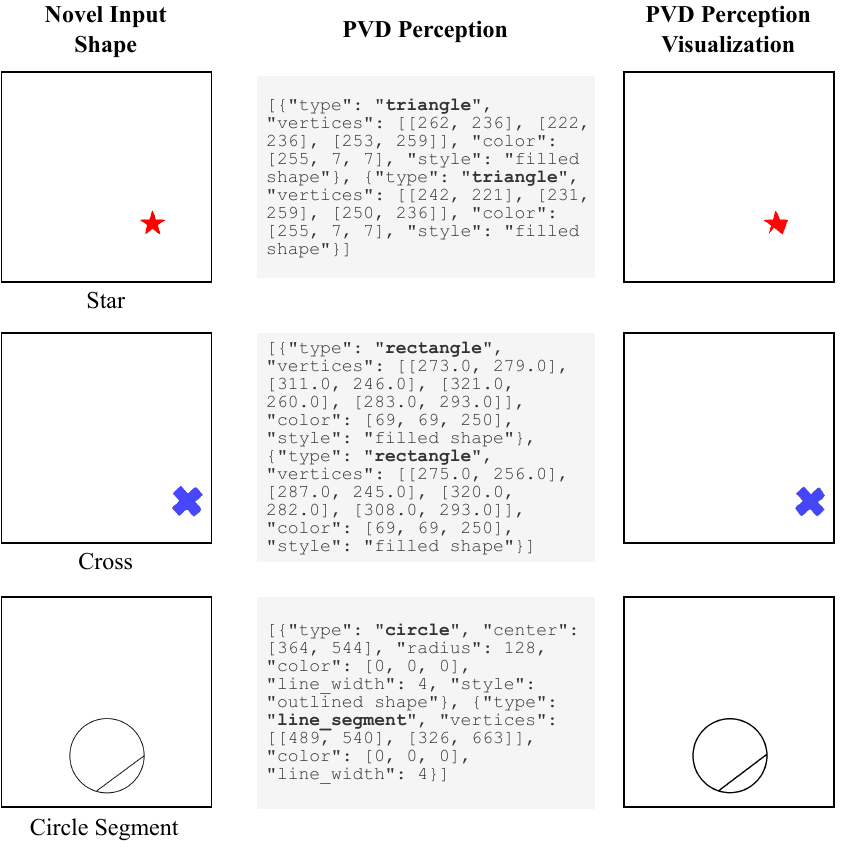}
  \vspace{-5pt}
  \caption{Qualitative examples on PVD model approximating novel concepts.}
  \label{fig:novel_concept_parsing}
\end{figure*}

\section{Additional Qualitative Examples on PVD Parsing Novel Concepts}
\label{app:tmlr_added_pvd_parsing_novel_concept}
We highlight that our PVD ontology constitutes a minimal but powerful set of primitives for expressing shapes in vector graphics. We demonstrate that the current PVD model shows promise in approximating novel shapes through composition. In Figure~\ref{fig:novel_concept_parsing}, we show several examples of the PVD model parsing novel concepts such as “star,” “cross,” and “circle segment.”

\vspace{-5pt}
\begin{table*}[ht]
\centering
\resizebox{0.68\textwidth}{!}
{
\centering

\begin{tabular}{@{}lll|ccc}
    \toprule
    \textbf{\begin{tabular}[c]{@{}c@{}} Prompt Version \end{tabular}} &
    \textbf{\begin{tabular}[c]{@{}c@{}} Reasoner \end{tabular}} & 
    \textbf{\begin{tabular}[c]{@{}c@{}} PVD Model \end{tabular}} &
    \textbf{\begin{tabular}[c]{@{}c@{}} SW-S\\2Obj \end{tabular}} &
    \textbf{\begin{tabular}[c]{@{}c@{}} SW-S\\mObj \end{tabular}}&
    \textbf{\begin{tabular}[c]{@{}c@{}} SW\\Sup \end{tabular}} \\
    \midrule
    - & GPT-4o & - &  0.97 & 0.81 & 0.92 \\
    Original & GPT-4o & Mistral-7B & 0.91 & 0.82 & 0.82 \\
    New & GPT-4o & Mistral-7B & 0.98 & 0.81 & 0.88 \\
    \bottomrule
\end{tabular}
}
\vspace{-5pt}
\caption{ShapeWorld task performance with new prompt.}
\vspace{-10pt}
\label{tab:prompt_engineering}
\end{table*}

\section{Further Exploration in Prompt Engineering}
\label{app:tmlr_added_prompt_engineering}
Observing that the PVD representation can sometimes be inaccurate and may degrade performance (i.e., on ShapeWorld tasks), we explore introducing a verification step in the prompt. Specifically, we instruct the model to first “double-check whether the objects in the PVD perception match the objects in the image.” 
We find that this step can enhance performance on tasks where the initial image-only baseline is already reasonably strong. Table~\ref{tab:prompt_engineering} shows the performance comparison on ShapeWorld tasks with the same LMM reasoner and PVD representation.


\definecolor{percepblue}{HTML}{0000CC}
\definecolor{taskorange}{HTML}{FF8000}

\section{Full Response of the Example in Figure~\ref{fig:overview}}
\label{sec_app:full_response_of_overview_example}
\begin{figure*}[t]
  \centering
  \includegraphics[width=\textwidth]{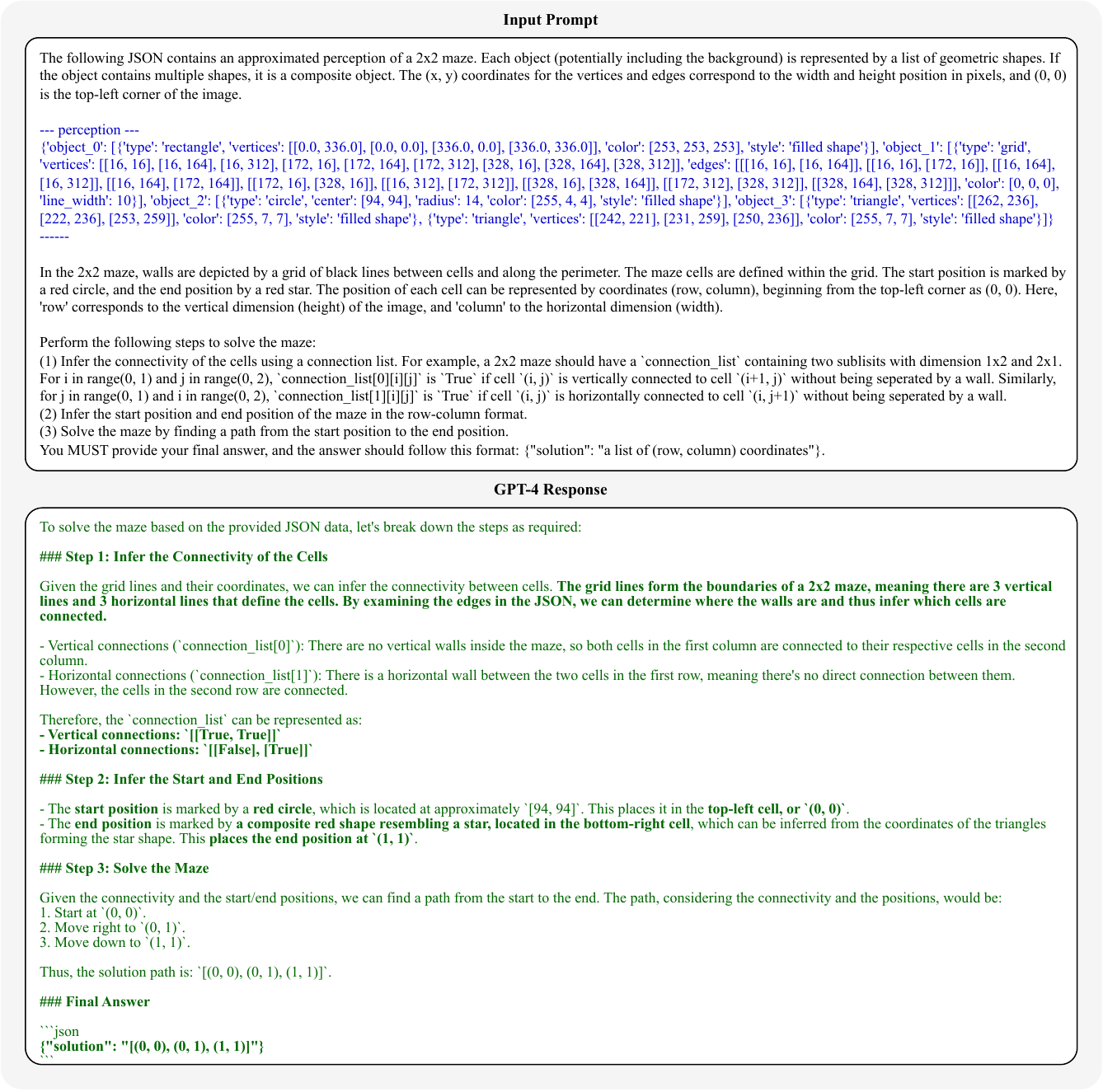}
  \caption{Full input prompt and GPT-4 response of the 2x2 maze solving example in Figure~\ref{fig:overview}. The \textcolor{percepblue}{blue} part in the input prompt indicates the generated \bl{} (\blabbr{}) of the entire image.}
  \label{fig:full_response_of_maze_example}
\end{figure*}

See Figure~\ref{fig:full_response_of_maze_example} for the full input prompt and the generated response from GPT-4 on the 2$\times$2 maze-solving task shown in Figure~\ref{fig:overview}.

\section{Task Prompts}
\label{sec_app:full_prompts}

\begin{figure*}[t]
  \centering
  \includegraphics[width=\textwidth]{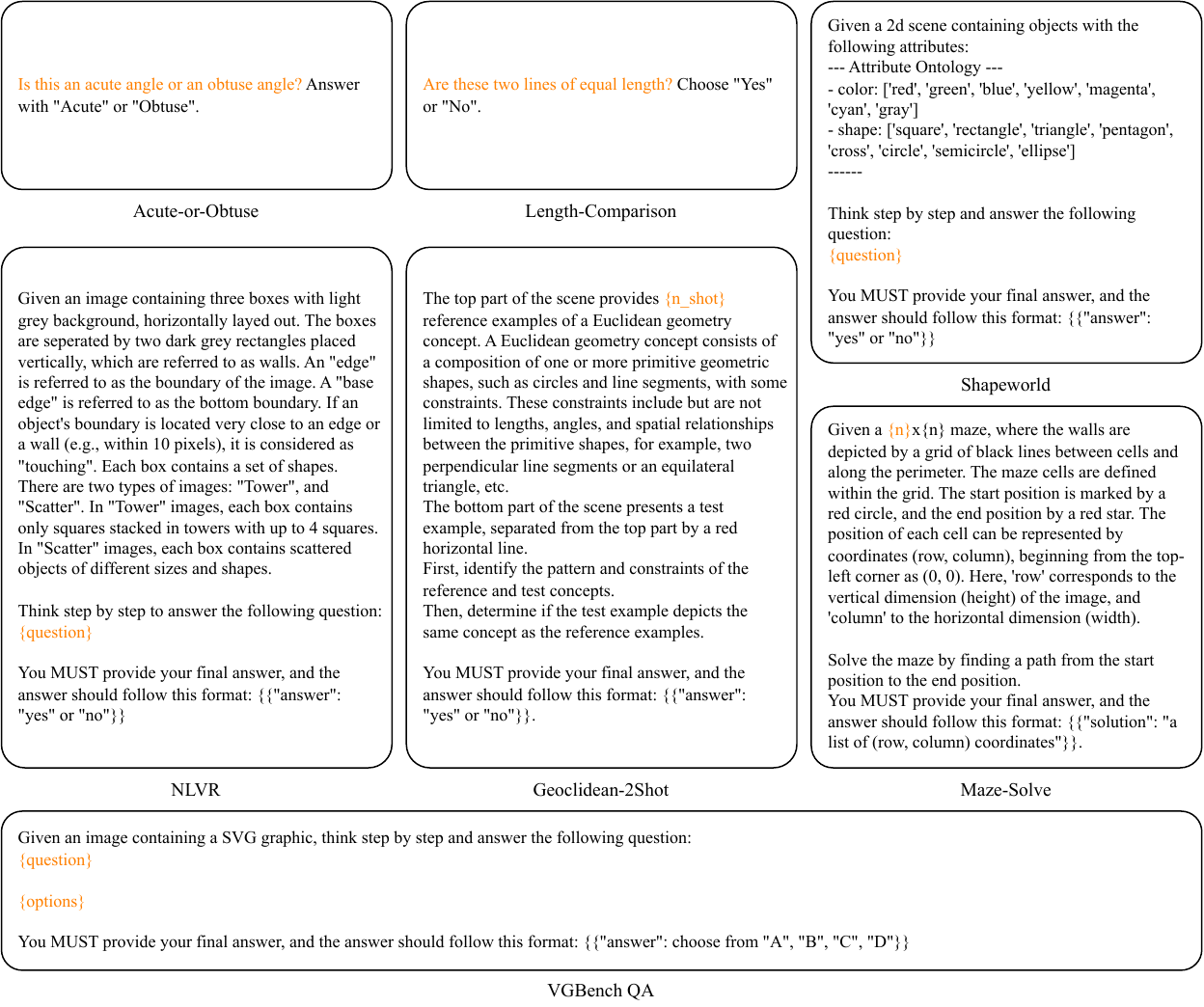}
  \caption{Prompts for zero-shot downstream tasks with image input}
  \label{fig:image_input_prompts}
\end{figure*}

Figure~\ref{fig:image_input_prompts} shows the prompts for models with only image representations as visual inputs.

\begin{figure*}[t]
  \centering
  \includegraphics[width=\textwidth]{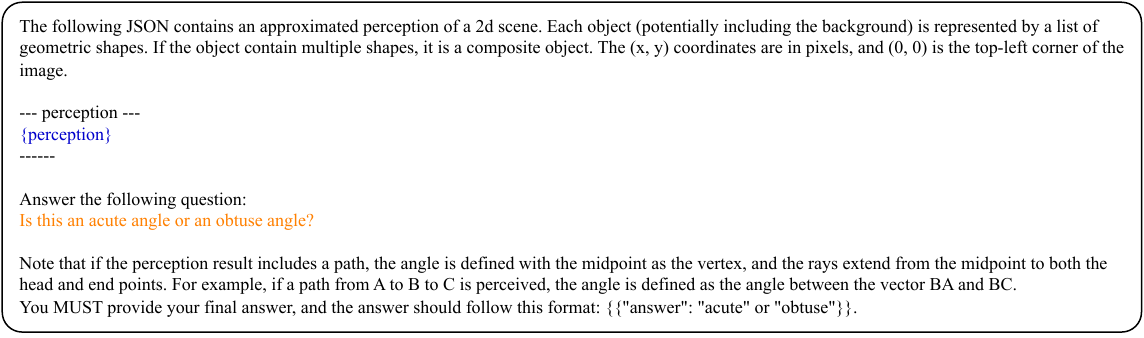}
  \caption{Prompt for task \dtaoo with \bl{} perception input.}
  \label{fig:bl_prompt_aoo}
\end{figure*}

\begin{figure*}[t]
  \centering
  \includegraphics[width=\textwidth]{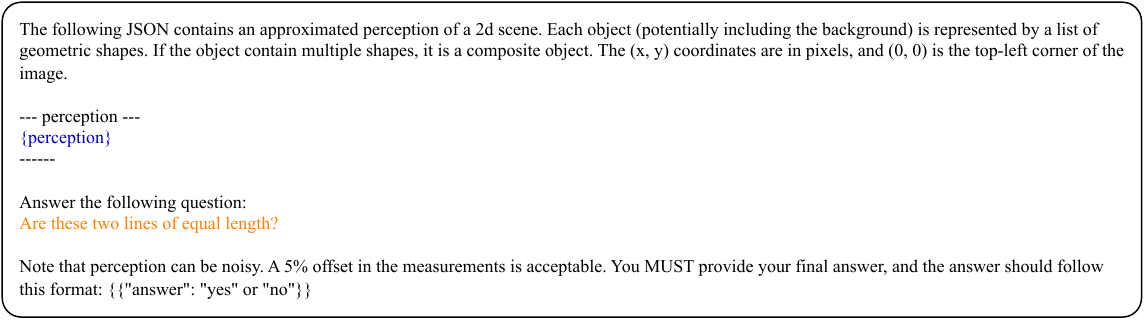}
  \caption{Prompt for task \dtlc with \bl{} perception input.}
  \label{fig:bl_prompt_lc}
\end{figure*}

\begin{figure*}[t]
  \centering
  \includegraphics[width=\textwidth]{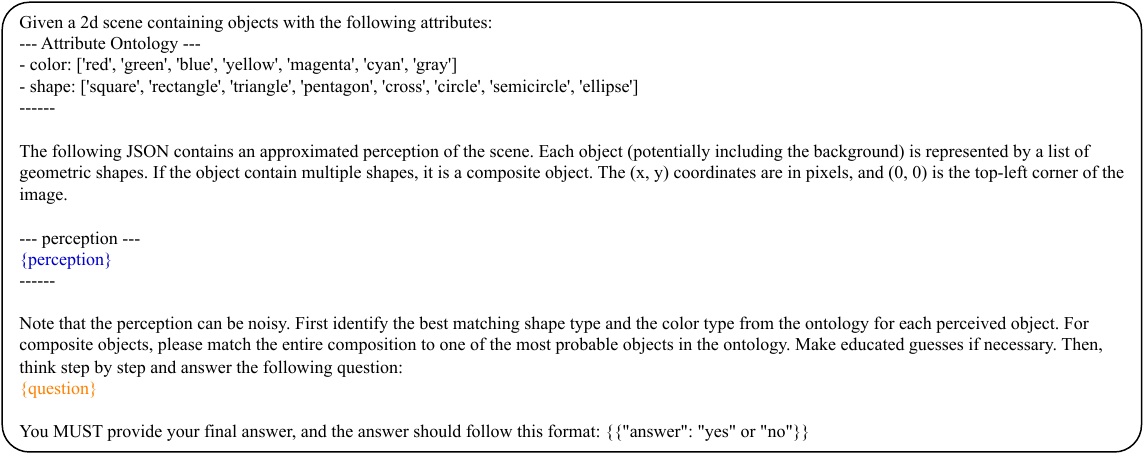}
  \caption{Prompt for task \dtswtobj with \bl{} perception input.}
  \label{fig:bl_prompt_swtobj}
\end{figure*}

\begin{figure*}[t]
  \centering
  \includegraphics[width=\textwidth]{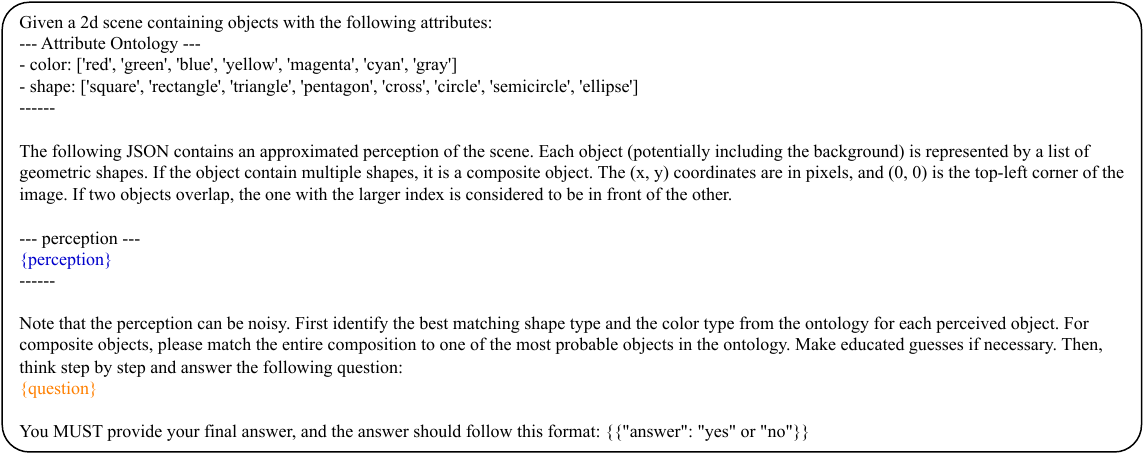}
  \caption{Prompt for task \dtswmobj with \bl{} perception input.}
  \label{fig:bl_prompt_swmobj}
\end{figure*}

\begin{figure*}[t]
  \centering
  \includegraphics[width=\textwidth]{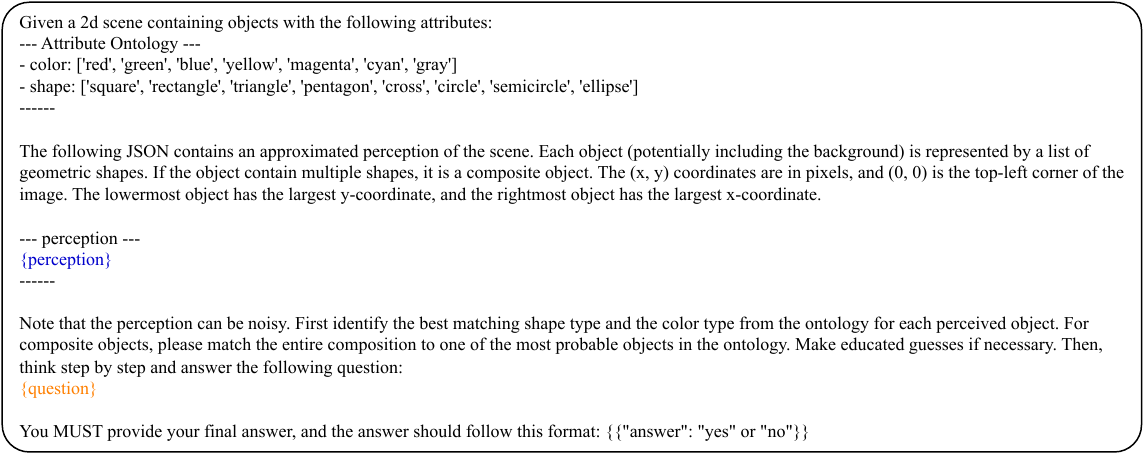}
  \caption{Prompt for task \dtswsup with \bl{} perception input.}
  \label{fig:bl_prompt_swsup}
\end{figure*}

\begin{figure*}[t]
  \centering
  \includegraphics[width=\textwidth]{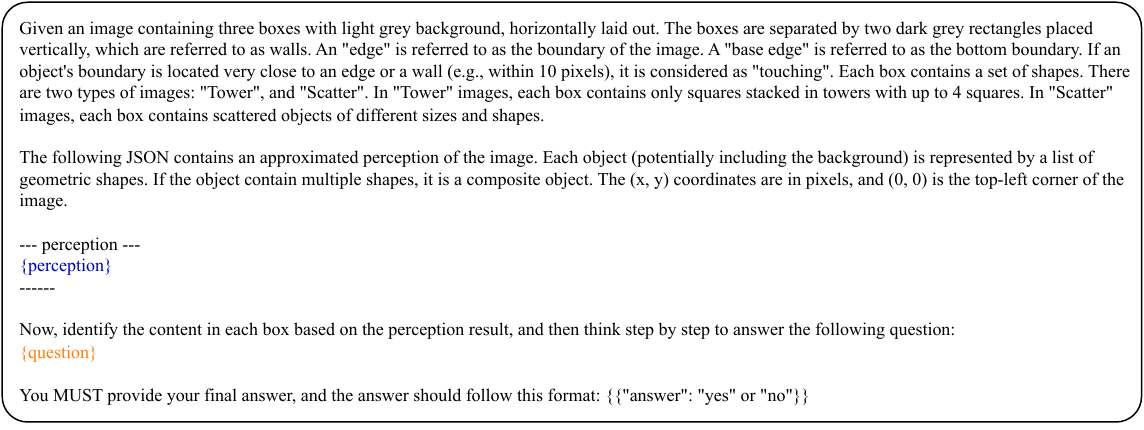}
  \caption{Prompt for task \dtnlvr with \bl{} perception input.}
  \label{fig:bl_prompt_nlvr}
\end{figure*}

\begin{figure*}[t]
  \centering
  \includegraphics[width=\textwidth]{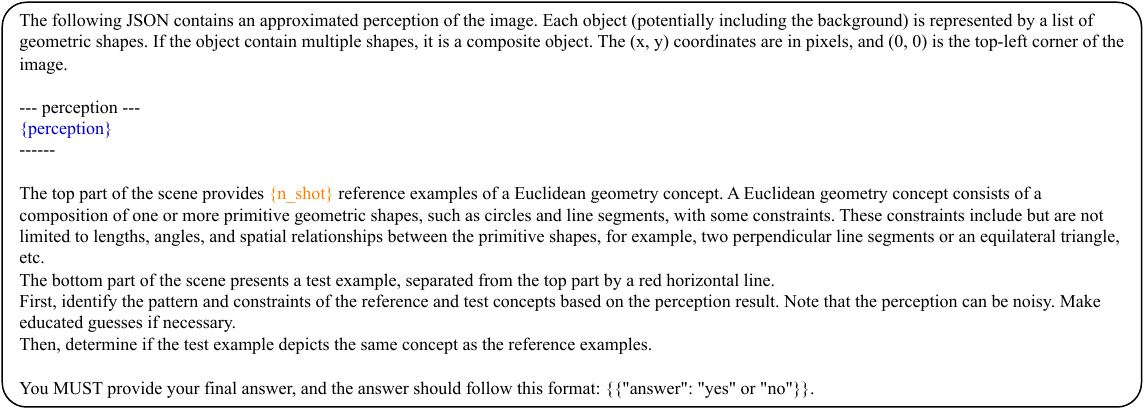}
  \caption{Prompt for task \dtgeo with \bl{} perception input.}
  \label{fig:bl_prompt_geoclidean}
\end{figure*}

\begin{figure*}[t]
  \centering
  \includegraphics[width=\textwidth]{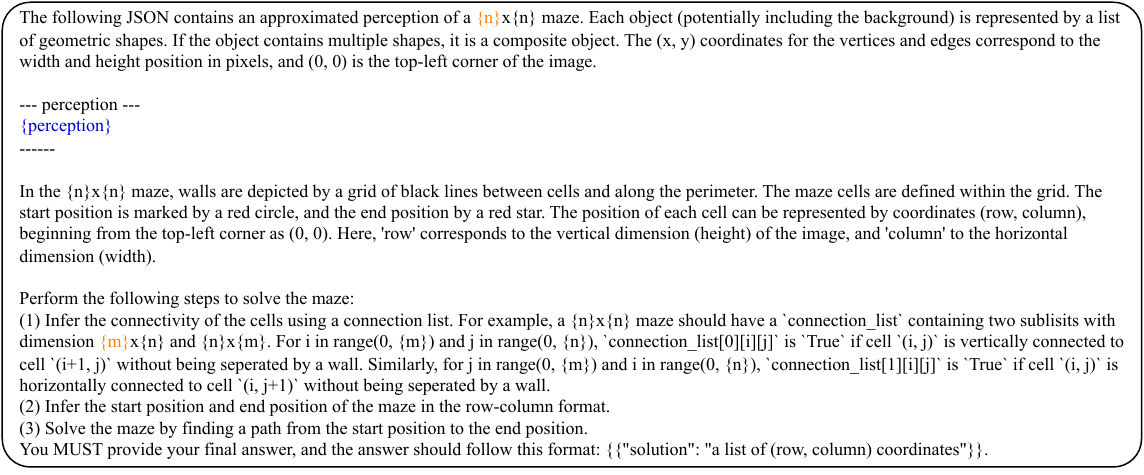}
  \caption{Prompt for task \dtmaze with \bl{} perception input.}
  \label{fig:bl_prompt_maze}
\end{figure*}

\begin{figure*}[t]
  \centering
  \includegraphics[width=\textwidth]{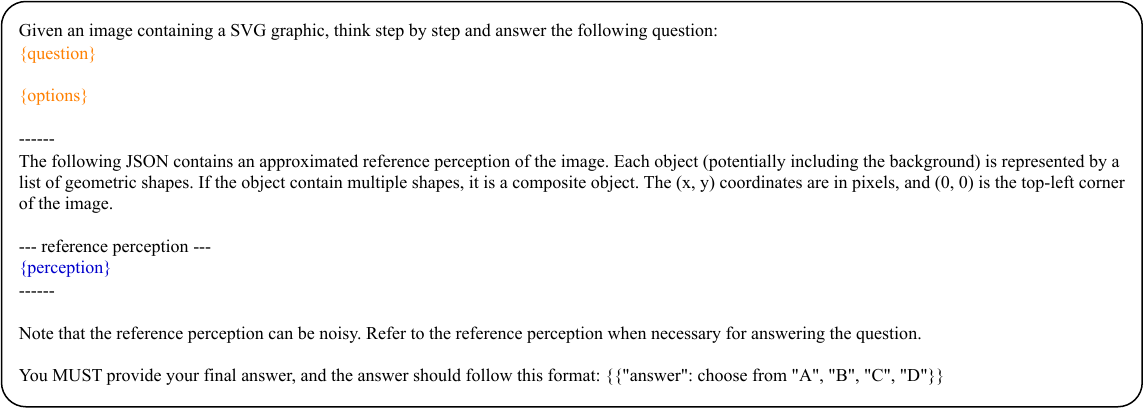}
  \caption{Prompt for \dtvgbench tasks with \bl{} perception input.}
  \label{fig:bl_prompt_vgbench}
\end{figure*}

Figures~\ref{fig:bl_prompt_aoo}-\ref{fig:bl_prompt_vgbench} show the prompts for \oursfull{}, where \textcolor{percepblue}{\{perception\}} will be filled with the aggregated \bl{} perception result, and the \textcolor{taskorange}{orange text} are instance-specific inputs such as the question. 
For \oursmm, the original image input will be preserved and feed to the LMM reasoner along with the filled prompt. 
Since the reasoning in \ourstxt is based solely on the \blabbr{} representation which is purely textual, task instructions that assume visual inputs can become ambiguous. For example, in the task \dtaoo, it is unclear which angle the question is referring to if we are only given the coordinates of two undirected edges. Therefore, we design task-specific prompts that remove such ambiguity.
Another noteworthy point is that, in contrast to visual inputs that naturally accommodate a degree of imprecision, symbolic representations lack such inherent leniency. For instance, even if two line segments differ by only one pixel in length, they might be considered identical in visual representations, but symbolic representations would likely identify them as different. To reintroduce a level of tolerance in tasks that involve arithmetic reasoning, such as length comparison, we incorporate task-specific instructions to account for a reasonable margin of error, like 5\%.

\section{Newly Constructed Downstream Task Datasets}
\label{sec_app:new_downstream_tasks}

\paragraph{\dtaoo{}.}
We use the Geoclidean data generator\footnote{\url{https://github.com/joyhsu0504/geoclidean_framework}} to generate images containing a single acute or obtuse angle with randomized orientations and ray lengths. The domain-specific language for generating the two concepts are shown as follows:
\begin{itemize}
    \item Acute Angle:
        \begin{center}
        \vspace{-3pt}
        \begin{tcolorbox}[colback=bggray, 
                          boxrule=0pt, 
                          arc=1.5mm,
                          width=0.45\textwidth, 
                          center title,
                          left=2mm, 
                          right=2mm, 
                          top=1mm, 
                          bottom=1mm] 
        \small
                \texttt{"l1* = line(p1(), p2())",}\\
                \texttt{"c1* = circle(p1(), p2())",}\\
                \texttt{"c2* = circle(p2(), p1())",}\\
                \texttt{"l2* = line(p3(c1, c2), p4(c1, c2))",}\\
                \texttt{"l4 = line(p5(l1, l2), p7(l1))",}\\
                \texttt{"l5 = line(p6(l2), p7(l1))"}
        \end{tcolorbox}
        \vspace{-3pt}
        \end{center}

\item Obtuse Angle:
        \begin{center}
        \vspace{-3pt}
        \begin{tcolorbox}[colback=bggray, 
                          boxrule=0pt, 
                          arc=1.5mm,
                          width=0.45\textwidth, 
                          center title,
                          left=2mm, 
                          right=2mm, 
                          top=1mm, 
                          bottom=1mm] 
        \small
                \texttt{"l1* = line(p1(), p2())",}\\
                \texttt{"c1* = circle(p1(), p2())",}\\
                \texttt{"c2* = circle(p2(), p1())",}\\
                \texttt{"l2* = line(p3(c1, c2), p4(c1, c2))",}\\
                \texttt{"l3* = line(p5(l1, l2), p6(l2))",}\\
                \texttt{"l4* = line(p5(l1, l2), p7(l1))",}\\
                \texttt{"l5* = line(p6(l2), p7(l1))",}\\
                \texttt{"l6* = line(p8(l3, l4), p9(l5))",}\\
                \texttt{"l100* = line(p5(c1, c2), p10(l6))",}\\
                \texttt{"c101* = circle(p5(c1, c2), p10(l6))",}\\
                \texttt{"c102* = circle(p10(l6), p5(c1, c2))",}\\
                \texttt{"l101* = line(p100(c101, c102), p101(c101, c102))",}\\
                \texttt{"l7 = line(p11(l100, l101), p6(l2))",}\\
                \texttt{"l8 = line(p11(l100, 101), p7(l1))"}
        \end{tcolorbox}
        \vspace{-3pt}
        \end{center}

\end{itemize}

\paragraph{\dtlc{}.} We use matplotlib\footnote{\url{https://matplotlib.org/stable/}} to plot two non-intersecting line segments on a canvas. 
These line segments may either be of identical length or of differing lengths. In scenarios where the lengths vary, we ensure the discrepancy is substantial (exceeding 15\% relative to the length of the shorter line segment) to ensure perceptibility.
The orientation of each line segment is determined independently and randomly, being either horizontal or vertical.

\paragraph{\dtmaze{}.} We leverage the maze-dataset package\footnote{\url{https://github.com/understanding-search/maze-dataset/tree/main}} to generate 2D unsolved mazes along with their corresponding ground truth solutions. We use "circle" shape to denote the start position and "star" shape to denote the end position. We generate two subsets featuring 2$\times$2 and 3$\times$3 maze configurations.

\section{Dataset Statistics}
\label{sec_app:dataset_stat}
\begin{table*}[th]
\centering
\resizebox{\textwidth}{!}
{
\centering

\begin{tabular}{@{}llcc}
    \toprule 
    & & \textbf{\begin{tabular}[c]{@{}c@{}} \# Training Instances \end{tabular}} &
        \textbf{\begin{tabular}[c]{@{}c@{}} \# Eval Instances \end{tabular}} \\
    \midrule
    \multirow{6}{*}{\textbf{\begin{tabular}[c]{@{}l@{}} Probing Tasks \end{tabular}}} & \ptloa{} & 10K & 1K\\
        & \ptaoo{} & 10K & 1000 \\
        & \ptlc{} & 10K & 1000 \\
        & \ptcq{} & 36K & 1000 \\
        & \ptsw{} & 15K & 100\\
        & \ptmaze{} & 10K & 600 \\
    \midrule
    \multirow{9}{*}{\textbf{\begin{tabular}[c]{@{}l@{}} Zero-Shot\\Downstream Tasks \end{tabular}}} & \dtaoo{} & - & 100 \\
        & \dtlc{} & - & 100 \\
        & \dtswtobj{} & - & 100 \\
        & \dtswmobj{} & - & 100 \\
        & \dtswsup{} & - & 100 \\
        & \dtnlvr{} & - & 100 \\
        & \dtgeo{} & - & 100 \\
        & \dtmazetwo{} & - & 100 \\
        & \dtmazethree{} & - & 100 \\
        & \dtvgbenchcat{} & - & 100 \\
        & \dtvgbenchcolor{} & - & 100 \\
        & \dtvgbenchusage{} & - & 100 \\
    \bottomrule
\end{tabular}
}
\caption{Statistics of the probing tasks (\S~\ref{subsec_app:svg_probing_exp}) and the downstream tasks (\S~\ref{sec:results}). The GPT-4(V) zero-shot results on probing tasks are reported on 100 randomly sub-sampled instances from the entire eval split. }

\label{tab:dataset_stats}
\end{table*}
Detailed statistics of the probing tasks used in \S~\ref{sec_app:preliminary_exp} and the zero-shot downstream tasks mentioned in \S~\ref{sec:results} can be found in Table~\ref{tab:dataset_stats}.


\end{document}